\documentclass{article}

\usepackage{arxiv}

\usepackage[utf8]{inputenc} % allow utf-8 input
\usepackage[T1]{fontenc}    % use 8-bit T1 fonts
\usepackage{hyperref}       % hyperlinks
\usepackage{url}            % simple URL typesetting
\usepackage{booktabs}       % professional-quality tables
\usepackage{amsfonts}       % blackboard math symbols
\usepackage{nicefrac}       % compact symbols for 1/2, etc.
\usepackage{microtype}      % microtypography
\usepackage{lipsum}
\usepackage{graphicx}
\graphicspath{ {./images/} }
\usepackage[authoryear,longnamesfirst]{natbib}
\usepackage{amssymb}
\usepackage{amsmath}
\usepackage{hyperref}
\usepackage{times,latexsym}
\usepackage{url}
\usepackage[T1]{fontenc}
\usepackage{graphicx} 
\usepackage{multirow}
\usepackage{xspace,mfirstuc,tabulary}
\usepackage{array}
\usepackage{makecell}
\usepackage{tabularx}
\usepackage{booktabs}
\usepackage{ragged2e}
\usepackage{pdflscape}
\usepackage{rotating}
\usepackage[inline]{enumitem}
\usepackage{hhline}
\usepackage{caption}
\usepackage{tikz}
\usetikzlibrary{tikzmark,calc}
\usepackage{pifont}
\usepackage{comment}
\usepackage[table]{xcolor}
\usepackage{arydshln}

\newcolumntype{C}[1]{>{\centering\arraybackslash}p{#1}}

\newcommand{\frameworkName}{\textsc{YARN }}
\newcommand{\frameworkNameDetail}{\textsc{Yielding Abstractions for Reasoning in Narratives}} 

\definecolor{LightPurple}{HTML}{d1afdb}
\definecolor{LightBlue}{HTML}{50d3d8}
\definecolor{lightgrayrow}{HTML}{d5dadb}
\definecolor{lightbluerow}{HTML}{c5fdff}

\title{Enhancing Structural Mapping with LLM-derived Abstractions for Analogical Reasoning in Narratives}

\author{
 Mohammadhossein Khojasteh \\
  Department of Computer Science\\
  Vrije Universiteit Amsterdam\\
  \texttt{m.khojasteh@vu.nl} \\
   \And
 Yifan Jiang \\
  Information Sciences Institute\\
  University of Southern California\\
  \texttt{yifjia@isi.edu} \\
  \And
 Stefano De Giorgis \\
  Department of Computer Science\\
  Vrije Universiteit Amsterdam\\
  \texttt{s.degiorgis@vu.nl} \\
  \And
 Frank van Harmelen \\
  Department of Computer Science\\
  Vrije Universiteit Amsterdam\\
  \texttt{frank.van.harmelen@vu.nl} \\
  \And
 Filip Ilievski \\
  Department of Computer Science\\
  Vrije Universiteit Amsterdam\\
  \texttt{f.ilievski@vu.nl} \\
}

\begin{document}
\maketitle
\begin{abstract}
Analogical reasoning is a key driver of human generalization in problem-solving and argumentation. Yet, analogies between narrative structures remain challenging for machines. Cognitive engines for structural mapping are not directly applicable, as they assume pre-extracted entities, whereas LLMs' performance is sensitive to prompt format and the degree of surface similarity between narratives. This gap motivates a key question: What is the impact of enhancing structural mapping with LLM-derived abstractions on their analogical reasoning ability in narratives?
To that end, we propose a modular framework named \frameworkName (\frameworkNameDetail), which uses LLMs to decompose narratives into units, abstract these units, and then passes them to a mapping component that aligns elements across stories to perform analogical reasoning. We define and operationalize four levels of abstraction that capture both the general meaning of units and their roles in the story, grounded in prior work on framing. Our experiments reveal that abstractions consistently improve model performance, resulting in competitive or better performance than end-to-end LLM baselines. Closer error analysis reveals the remaining challenges in abstraction at the right level, in incorporating implicit causality, and an emerging categorization of analogical patterns in narratives. \frameworkName enables systematic variation of experimental settings to analyze component contributions, and to support future work, we make the code for \frameworkName openly available.\footnote{Code available here: https://github.com/mhkhojaste/narrative-analogy}
\end{abstract}

% keywords can be removed

\setcounter{footnote}{1}
\section{Introduction}

% \textbf{Background}
Analogical reasoning is the process of identifying a common structure between two situations or domains and using that structure to draw further conclusions or insights \citep{GENTNER2012130}. 
Analogy is a key driver of human generalization across tasks ranging from simple comparisons to highly complex reasoning. Analogy enables humans to interpret or explain unfamiliar concepts by mapping them onto familiar ones \citep{holyoak1996mental}. As such, it plays a central role in complex domains and tasks: legal argumentation, business decision-making, causal explanation, creative problem-solving \citep{penn2008darwin}, and even non-traditional forms of reasoning, such as interpreting poetic metaphors \citep{penn2008darwin}. Due to its wide usage, from learning to creativity, analogical skills have been considered a core process of human cognition \citep{GENTNER2012130, penn2008darwin, hofstadter2001analogy}. 

\begin{figure*} 
    \centering 
    \includegraphics[width=\linewidth]{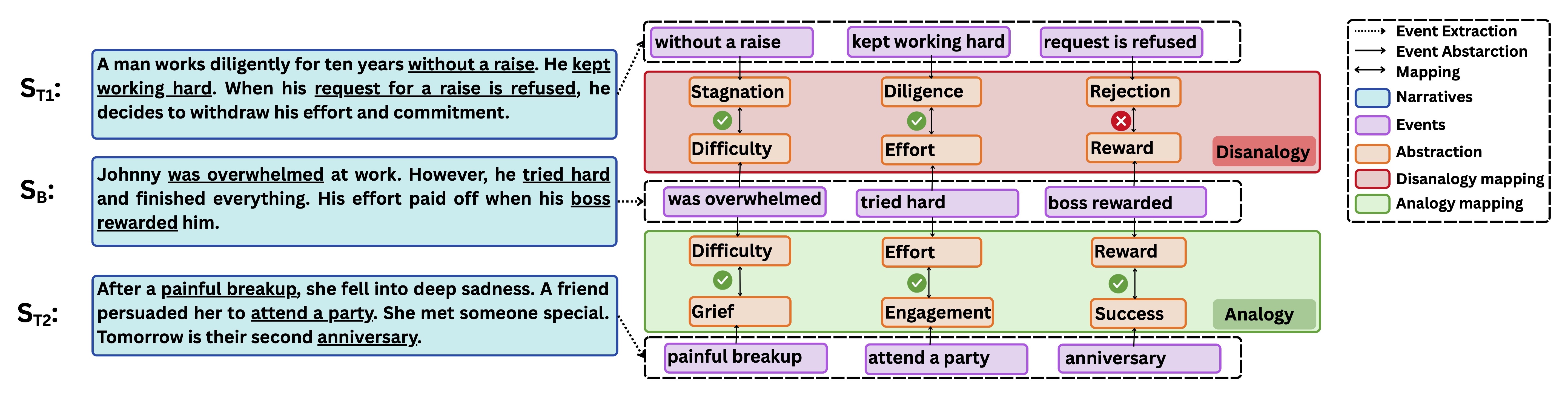} 
    \caption{Structural mapping example derived from ARN dataset narratives \citep{sourati-etal-2024-arn}. The \textcolor{LightBlue}{narratives} are decomposed into \textcolor{LightPurple}{units}, which are then converted into \textcolor{orange}{abstractions} to capture their roles and general meaning, thus facilitating a structural mapping. $S_B$ and $S_{T2}$ form an \textcolor[RGB]{0,100,0}{analogy}, whereas $S_B$ and $S_{T1}$ are \textcolor{red}{disanalogous} because their final events are opposed to each other. In this example, the system needs to prioritize far analogy over a near disanalogy, i.e., relational over surface similarity.}
    \label{fig:example} 
\end{figure*}

Narratives emerge as a natural format for estimating the ability of machines or humans to perform analogies between two domains \citep{sourati-etal-2024-arn, jiayang-etal-2023-storyanalogy}. We define a \textit{narrative} as a coherent sequence of events shaped by a particular point of view \citep{popova2014narrativity}. Narratives are well-suited for studying analogy because they naturally organize events into roles and dependencies, and analogy is used to map the event's functional and causal organization from one story onto another (i.e., structure similarity), rather than by matching lexical overlap or topical similarity (i.e., surface similarity) \citep{sourati-etal-2024-arn}. \autoref{fig:example} illustrates a base narrative and two candidate target narratives, where the goal is to decide which of the candidate targets is analogous to the base. Deciding on the analogy from the narratives directly is challenging. However, once we extract events and abstract them according to their meaning in the corresponding story, it becomes apparent that $S_{T1}$ is a disanalogy (as shown by the mismatch between Rejection and Reward) and $S_{T2}$ is an analogy to $S_B$ (as all abstractions map logically between these two narratives).

% \textbf{Research gap}

% Modeling and evaluating analogical reasoning in AI systems has become an important and increasingly active line of research \citep{sourati-etal-2024-arn, jacob-etal-2023-fame}. Prior work has introduced tasks and benchmarks aimed at measuring whether models can perform analogical reasoning in ways that resemble human performance \citep{lewis2024using, stevenson2024can}. One setting for this evaluation is narratives \citep{sourati-etal-2024-arn, jiayang-etal-2023-storyanalogy}. 

How can a machine draw such conclusions about analogies between narratives? Cognitive theories and their corresponding implementations \citep{gentner1983structure,holyoak1989analogical,hofstadter1995copycat} draw inspiration from the way humans form analogies. A particularly popular theory is the Structural Mapping Theory (SMT) \citep{gentner1983structure}, which describes analogical reasoning as the alignment of relational structures between a base and a target. The Structural Mapping Engine (SME) \citep{falkenhainer1989structure} operationalizes SMT to map between structured representations. This limitation of SME and related engines to operate on unstructured data has been recognized, leading to methods that complement SME to map entities using automatically extracted relations \citep{jacob-etal-2023-fame} or with emotional taxonomies \citep{thagard2001emotional}. However, cognitive engines for structural mapping still assume pre-extracted entities and are thus not directly applicable to unstructured data such as narratives.

Studies of analogical reasoning in AI and NLP have predominantly focused on proportional analogies (e.g., King:Queen :: Man:Woman). Typical approaches employ static word embeddings that model relationships as geometric regularities in vector space \citep{mikolov2013efficient} or contextual embeddings from pretrained language models \citep{ushio2021bert}. Recently, Large Language Models (LLMs) have been reported to exhibit emergent analogy-solving behavior \citep{webb2023emergent}, although follow-up analyses highlighted substantial limitations, suggesting that these capabilities probably stem from data contamination or surface pattern matching rather than from analogical reasoning abilities reminiscent of those of humans \citep{lewis2024using, stevenson2024can}. Prior studies identify two underlying issues: difficulty in capturing the generalized meaning of units \citep{opielka2025analogical}, which correspond to the abstractions in \autoref{fig:example}, and limitations in understanding and reliably transferring causal relations across contexts \citep{lee2025curious}. Analogies between narratives, as in \autoref{fig:example}, have only recently gained interest in AI and NLP, revealing human-level performance on near analogies (high surface similarity), but low, and often below-random, performance on far analogies (across two domains with low surface similarity) \citep{sourati-etal-2024-arn, jiayang-etal-2023-storyanalogy}. Samples such as those in \autoref{fig:example} are particularly challenging, as the system needs to prioritize a far analogy over a near disanalogy, i.e., relational over surface similarity.

We identify a gap between cognitive frameworks such as SMT and LLMs for analogical reasoning in narratives, which we summarize as three challenges. First, existing \textit{cognitive frameworks assume structured inputs and therefore cannot be applied directly} to unstructured narratives without an additional step to build a representation.
Second, previous work does not provide \textit{conclusive evidence on what those representations should be or how they could be integrated with cognitive frameworks} for structural mapping.
Third, \textit{LLMs' performance is sensitive to prompt format \citep{errica2025did} and the degree of surface similarity} between narratives \citep{sourati-etal-2024-arn}. 
Intersecting these three challenges is the observation that the relationship between LLMs and structural mapping frameworks, both in terms of complementarity and integration, has not yet been systematically explored, as also noted in recent work \citep{gentner2025structure, petersen2025modelling}.

This gap motivates a natural question: \textit{What is the effect of enhancing structural mapping with LLM-derived abstractions on their analogical reasoning ability in narratives?} We devise \frameworkName\unskip,\footnote{Translation via Google: A long or \underline{rambling} story, especially one that is \underline{implausible}.} the first framework that combines LLM-based extraction with a structure mapping component for analogical reasoning in narratives (\autoref{fig:pipeline_overview}). Our experiments with this framework reveal key insights into the promises and challenges in this direction and point to a rich set of future directions for cognitively supported and LLM-enabled narrative analogy frameworks. 
Our contributions are:
\begin{enumerate*}
  \item \frameworkName \unskip, a \textbf{neurosymbolic framework} for systematic experimentation in analogical reasoning in narratives that connects representations extracted from LLM with structural mapping. We use an LLM to decompose narratives into units, abstract them, and then pass the abstracted units to a mapping component that aligns elements across stories to judge the analogical fit of the stories. % perform analogical reasoning.
  \item A \textbf{four-level abstraction} deployed in \frameworkName \unskip, tailored to narrative reasoning from a perspective of framing and pragmatic meaning of narrative events. These levels are designed to capture complementary aspects of the unit's meaning and the narrative function of events.
  \item \textbf{Systematic experiments} on two narrative analogy benchmarks, examining the impact of the components of our framework, with an emphasis on the four abstraction levels. We incorporate a detailed error analysis to distill the remaining challenges.
  \item A \textbf{detailed discussion} summarizes the limitations discovered to abstract at the right level, incorporates implicit causality, and categorizes emerging analogical patterns in narratives. We distill these limitations into open questions to be addressed in subsequent research.
\end{enumerate*}
Finally, we make the code for our \frameworkName experimental framework and data available to facilitate future research in these directions.

\section{Related Work}
\label{sec:related}
% We first review existing analogical reasoning models, beginning with cognitive theories and the engines derived from them, followed by approaches developed in classic NLP, and finally research on analogical reasoning with LLMs. We next examine story representation and abstraction, focusing on extracting representations for narrative text.

\subsection{Cognitive theories and computational frameworks for analogy} 

% Analogies vary in kind, ranging from attribute-based comparisons to proportional mappings between relational pairs, and to fully structural analogies that align systems of relations across domains \citep{turney2003combining, holyoak1996mental}. 
% While narratives have been used to study analogical reasoning \citep{gentner1986systematicity, gentner1983structure}, these studies have either focused on simplified settings or relied on human participants, and no computational systems capable of performing narrative analogies have been proposed \citep{lu2022probabilistic}.
% A central distinction in the literature contrasts surface mappings, which rely on matching object attributes, with relational or structural mappings, which align deeper relational structure between situations \citep{gentner1983structure}. This distinction closely relates to the contrast between near and far analogies: near analogies are supported by both surface and structural mapping and are therefore easier to recognize, whereas far analogies preserve relational structure despite minimal surface overlap, making them more cognitively demanding and more reliant on relational reasoning \citep{alexieva2017processing, gentner1983structure}.

Analogies span a wide spectrum, ranging from simple attribute-based comparisons to proportional mappings, and to fully structural analogies that align systems of relations across domains \citep{turney2003combining, holyoak1996mental}. Although narratives have been used to investigate analogical reasoning \citep{gentner1986systematicity, gentner1983structure}, such studies have focused mainly on simplified settings or human participants \citep{lu2022probabilistic}. A fundamental distinction in the literature contrasts surface mappings, which match object attributes, with structural mappings, which align deep relational structures \citep{gentner1983structure}. This distinction underpins the difference between near and far analogies: near analogies share both surface and structural features, making them easier to recognize, while far analogies preserve relational structure despite minimal surface overlap, requiring more cognitively demanding relational reasoning \citep{alexieva2017processing, gentner1983structure}.

Insights into human analogical reasoning have inspired a range of theoretical and computational frameworks. Gentner’s SMT posits that analogical reasoning involves mapping a base domain to a target domain by aligning their relational structures \citep{gentner1983structure}. This theory underpins the SME, a computational system that identifies one-to-one mappings between structured representations \citep{falkenhainer1989structure}, with recent extensions for greedy merging and incremental processing to handle new knowledge efficiently \citep{forbus2017extending}. A modern derivative, FAME (Flexible, Scalable Analogy Mappings Engine), adapts SME for entity mapping by combining LLM-based relation extraction with a greedy beam-search mapping algorithm \citep{jacob-etal-2023-fame}.
Alternative frameworks offer competing perspectives. ACME frames analogy as a parallel constraint satisfaction process governed by structural consistency, semantic similarity, and pragmatic constraints \citep{holyoak1989analogical}. Copycat models high-level analogies as emerging from small, low-level computational actions \citep{hofstadter1995copycat}, while LISA represents propositions as distributed patterns over semantic units to bind structured representations dynamically \citep{hummel1997distributed}.

A common limitation of these cognitive engines is their inability to process unstructured texts, such as narratives, directly. We bridge this gap by distilling unstructured text into structured, event-centric representations of narratives through a number of abstractions; these are subsequently fed into a structural mapping process that estimates the analogical fit via pairwise alignment.

\subsection{NLP models for analogical reasoning}

The treatment of analogical reasoning in NLP has evolved from static embeddings to generative models. Early work relied on the assumption that vector offsets in models like Word2Vec and GloVe could proxy for semantic relations \citep{mikolov2013efficient, pennington2014glove}. Later, this linear assumption was proven to be insufficient for robust reasoning \citep{rogers2017too}. Similar robustness challenges were discovered for contextualized embeddings \citep{ushio2021bert}.

The rise of LLMs reintroduced the possibility of computational analogy, with some claims of superhuman performance \citep{webb2023emergent}. However, these capabilities appear to be brittle: LLMs exhibit sharp performance drops when testing data shifts from their training distribution, failing to transfer abstract relations as humans do \citep{lewis2024using, stevenson2024can}. This deficiency is critical in the context of narrative analogies, where deep structural alignment is required. Current benchmarks such as ARN and StoryAnalogy show that LLMs failing to grasp far analogies \citep{jiayang-etal-2023-storyanalogy, sourati-etal-2024-arn}, which may be due to their struggles with abstraction and extraction of causal structures \citep{opielka2025analogical,lee2025curious}.

Existing research has largely restricted LLM evaluation to in-context learning on proportional analogies. In contrast, our neurosymbolic approach combines the generative power of LLMs with structural mapping: LLMs generate the representations, while the mapping component discovers the one-to-one correspondences required for valid analogical mapping.

\subsection{Story representation and abstraction}
\label{sec:story_rep}

While narrative analysis has a rich history~\citep{propp2010morphology}, LLMs are increasingly being used to extract structured, event-centric information from narratives. An event is typically formalized as a combination of a verb trigger, an event type, and associated participant roles \citep{liu2020extracting}. Recent systems like INCSCHEMA and NetPrompt employ incremental construction and layered prompting to build these event graphs \citep{li-etal-2023-open, mu2025netprompt}, while others leverage intermediate structures such as AMR for robust argument extraction \citep{hsu-etal-2023-ampere}.  Beyond individual events, recent work models inter-event relations, such as temporal ordering \citep{yuan-etal-2023-zero} and causal dependencies \citep{hu-etal-2025-large, sun-etal-2024-event}, as well as discourse-level features like narrative arcs and turning points \citep{tian-etal-2024-large-language}.

Despite these advances, extracted representations often remain tethered to surface forms. Narrative analogy requires \textit{abstraction}, broadly defined as decreasing specificity to increase scope \citep{gentner2017analogy}. However, abstraction is multifaceted \citep{barsalou2003abstraction, ilievski2025aligning}. A pragmatic lens is offered by Entman’s concept of framing, where specific elements are made salient to guide interpretation \citep{entman1993framing}. Sullivan et al. consolidate various notions of framing into three levels \cite{sullivan2023three}: semantic frames (lexical meaning structures) \citep{baker1998berkeley}; cognitive frames (scripts and schemas evoking situational expectations) \citep{minsky1974framework}; and communicative frames (word choice that steers interpretation) \citep{sullivan2023three}.

Although event-centric models dominate narrative representation \citep{li-etal-2023-open}, we note the lack of a unified conceptual and computational method for abstracting surface forms while preserving narrative structure. We address this by adapting Sullivan’s framework to a four-level abstraction, using LLMs to extract these hierarchical abstractions as the basis for structural analogical mapping.

\section{Method}
\label{method}

\begin{figure*}
    \centering 
    \includegraphics[width=\linewidth]{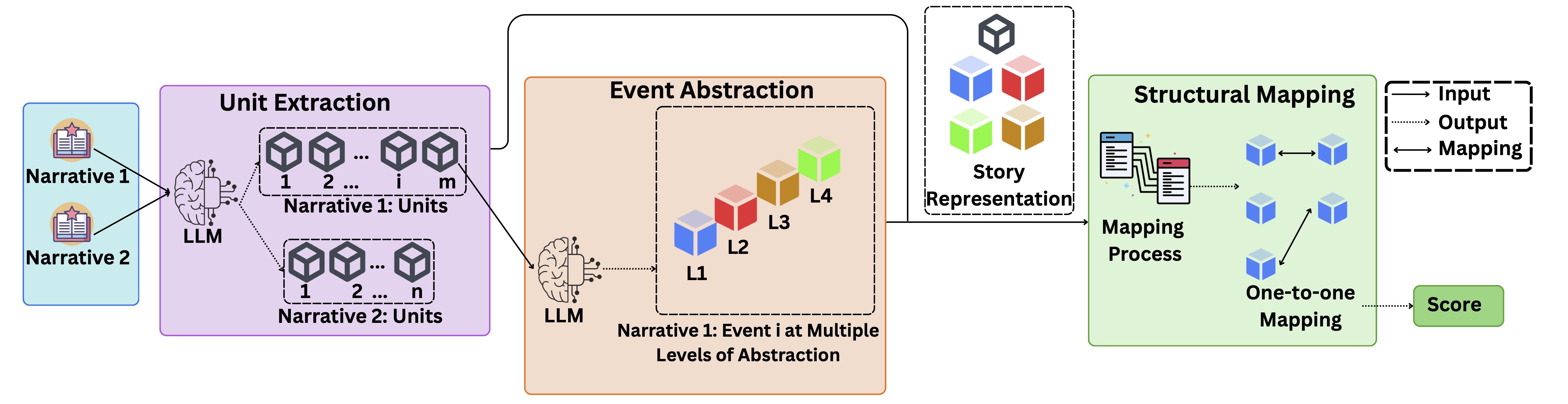} 
    \caption{A high-level overview of the \frameworkName pipeline: we first use LLMs to extract information by identifying units in the two narratives and converting them into abstractions, and then generate a structural one-to-one mapping between the units and abstractions of the two narratives.}
    \label{fig:pipeline_overview} 
\end{figure*}

Given a base story ($S_{base}$) and $n$ candidate target stories (${S_{target_1}, ..., S_{target_n}}$), our method \frameworkName aims to select the target that yields the most consistent \emph{structural} mapping to $S_{base}$.
It follows a three-phase pipeline (\autoref{fig:pipeline_overview}). First, \frameworkName decomposes each story into event-level units (\autoref{method:unit-extraction}). We define an event as a specific occurrence involving participants \citep{liu2020extracting}. For example, as shown in \autoref{tab:full-example}, \frameworkName extracts event phrases from the story, such as ``Johnny had too many projects''. Next, it converts each unit into multiple abstractions that capture its (a) general meaning, (b) narrative function (e.g., challenge, attempt, outcome), and (c) relations to other units, enabling alignment beyond surface differences (\autoref{method:abstraction}). For instance, ``Johnny had too many projects'' can be mapped to (a) ``tasks workload'', (b) ``struggle'', and (c) ``Main event''. \autoref{tab:full-example} presents a detailed example of extracted units and their conversion into different levels of abstraction. Then, \frameworkName considers all possible pairs of abstracted units across the two stories as candidate local mappings and score each pair using a similarity function. It combines these local mappings into a global one-to-one mapping with an overall score (\autoref{method:mapping}). This allows mappings such as ``tasks workload'' - ``relationship loss'', as both express a struggle and serve as the main event. This procedure yields scores for each candidate target relative to the base story and selects the highest-scoring mapping as the answer.

\subsection{Unit Extraction}
\label{method:unit-extraction}
Our \frameworkName framework converts a story $S$ into a set of units $U_s = \{U_1, U_2, ..., U_n\}$. Each unit $U_i$ is defined as a token span capturing a distinct and meaningful element of the story, forming the structural backbone for abstraction and mapping.
We impose two semantic constraints on the extracted units. i)~\textit{Coverage:} The units should preserve the story’s content with minimal information loss ($\bigcup_{i=1}^n U_i \approx S$). ii)~\textit{Non-redundancy:} Each unit should contribute unique information without overlap ($U_i \cap U_j = \varnothing, \forall i \neq j$). These constraints ensure a faithful and unambiguous representation for downstream structural mapping. 

We opt for event-centric representations of stories~\citep{jm3}, as events capture the core semantics of narratives while omitting some discourse cues (e.g., explicit connectives). Concretely, we use \textbf{event phrases}: short textual fragments describing single events and their arguments. This representation preserves the original wording while isolating minimal spans for structural analysis, producing phrases such as ``Johnny had too many projects''. Additionally, using event phrases combines the advantages of other representations: like full sentences, they preserve much of the textual context and like structured AMR-style representations \citep{wein2024survey}, they isolate individual events rather than grouping multiple events into a single unit. In addition to extracting events, \frameworkName also infers their temporal order, since textual order may differ from chronological order in narratives. These temporal indices help preserve structural consistency during abstraction. We use LLMs with few-shot prompting to extract event phrases. Details of the prompts for the extraction process are provided in \autoref{sec:setup}, and a summary of the event extraction prompt is provided in the appendix (\autoref{tab:event-prompt}).

\begin{table}[t]
\caption{The extracted event phrases and abstractions for the following story from the ARN benchmark \citep{sourati-etal-2024-arn}: ``Johnny had too many projects and too many short deadlines, and he was stressed. He kept working as hard as he could to finish everything. His boss noticed how hard he was working and offered him a raise.'' For each extracted event, we provide conceptual, evaluative, and narrative arc abstractions. Events that share the same narrative arc (indicated by dashed lines) are then grouped into a single stage abstraction. A detailed explanation of how each abstraction level is extracted is provided in \autoref{method:abstraction}.
}
\label{tab:full-example}
\scriptsize
\centering
\renewcommand{\arraystretch}{1.15}
\setlength{\tabcolsep}{4pt}

\begin{tabularx}{\linewidth}{
    >{\raggedright\arraybackslash}X
    >{\centering\arraybackslash}m{0.16\linewidth}
    >{\centering\arraybackslash}m{0.12\linewidth}
    >{\centering\arraybackslash}m{0.12\linewidth}
    >{\centering\arraybackslash}m{0.18\linewidth}
}

\hline
% \multicolumn{5}{c}{\textbf{Story}} \\
% \hline

% \multicolumn{5}{>{\raggedright\arraybackslash}p{\linewidth}}{
% Johnny had too many projects and too many short deadlines, and he was stressed. He kept working as hard as he could to finish everything. His boss noticed how hard he was working and offered him a raise.
% } \\
% \cdashline{1-5}

\textbf{Units} & 
\makecell[c]{\textbf{Conceptual}\\\textbf{abstraction}} & 
\makecell[c]{\textbf{Evaluative}\\\textbf{abstraction}} & 
\makecell[c]{\textbf{Narrative arc}\\\textbf{abstraction}} & 
\makecell[c]{\textbf{Stage}\\\textbf{abstraction}} \\
\hline

Johnny had too many projects 
& tasks workload 
& struggle 
& Main event 
& \multirow{2}{=}{\centering work overload}\\

Johnny had too many short deadlines 
& time pressure 
& struggle 
& Main event 
& \\
\cdashline{1-5}

Johnny was stressed 
& emotion stress 
& struggle 
& Challenge 
& stress from workload \\
\cdashline{1-5}

Johnny kept working hard 
& work dedication 
& effort 
& Action 
& \multirow{2}{=}{\centering \makecell[c]{persistent effort\\and completion}} \\

Johnny finished everything 
& tasks completion 
& gain 
& Action 
& \\
\cdashline{1-5}

Johnny's boss noticed his hard work 
& work recognition 
& gain 
& Conclusion 
& \multirow{2}{=}{\centering reward for dedication} \\

Johnny's boss offered a raise 
& work reward 
& gain 
& Conclusion 
& \\
\hline
\end{tabularx}

\end{table}

\subsection{Unit Abstraction}
\label{method:abstraction}
Event phrases alone are insufficient for structural mapping because they retain surface details (e.g., names and domain terms) and they represent events in isolation, without capturing their general meaning or roles within the story, which are essential for analogical reasoning \citep{GENTNER2012130}. \frameworkName therefore transforms each event phrase into an abstract representation (\textbf{abstraction}) that emphasizes relational structure. Given a story $S$ with event phrases $U_s = \{U_1, U_2, ..., U_n\}$, this step produces a set of abstractions $A_s = \{A_1, A_2, ..., A_m\}$, where $n \geq m$. Each abstraction expresses a more general meaning than its corresponding event phrase(s) and facilitates structural comparison between events.

Abstraction in narratives is inherently context-dependent, as the meaning of an event depends on its role within the narrative sequence \citep{gentner2017analogy, bolognesi2020abstraction, sullivan2023three}. The same event may admit different abstractions depending on context. For example, ``Tomorrow is their second anniversary'' in $S_{T2}$ in \autoref{fig:example} could correspond to abstractions such as ``anniversary'' or ``planning'' in isolation. However, in the context, this event follows a series of hardships; thus, a higher-level abstraction such as ``success'' or ``positive outcome'' captures its narrative function more precisely. This example illustrates that abstraction requires understanding event interactions and narrative structure.

\begin{figure*}
    \centering 
    \includegraphics[width=\linewidth]{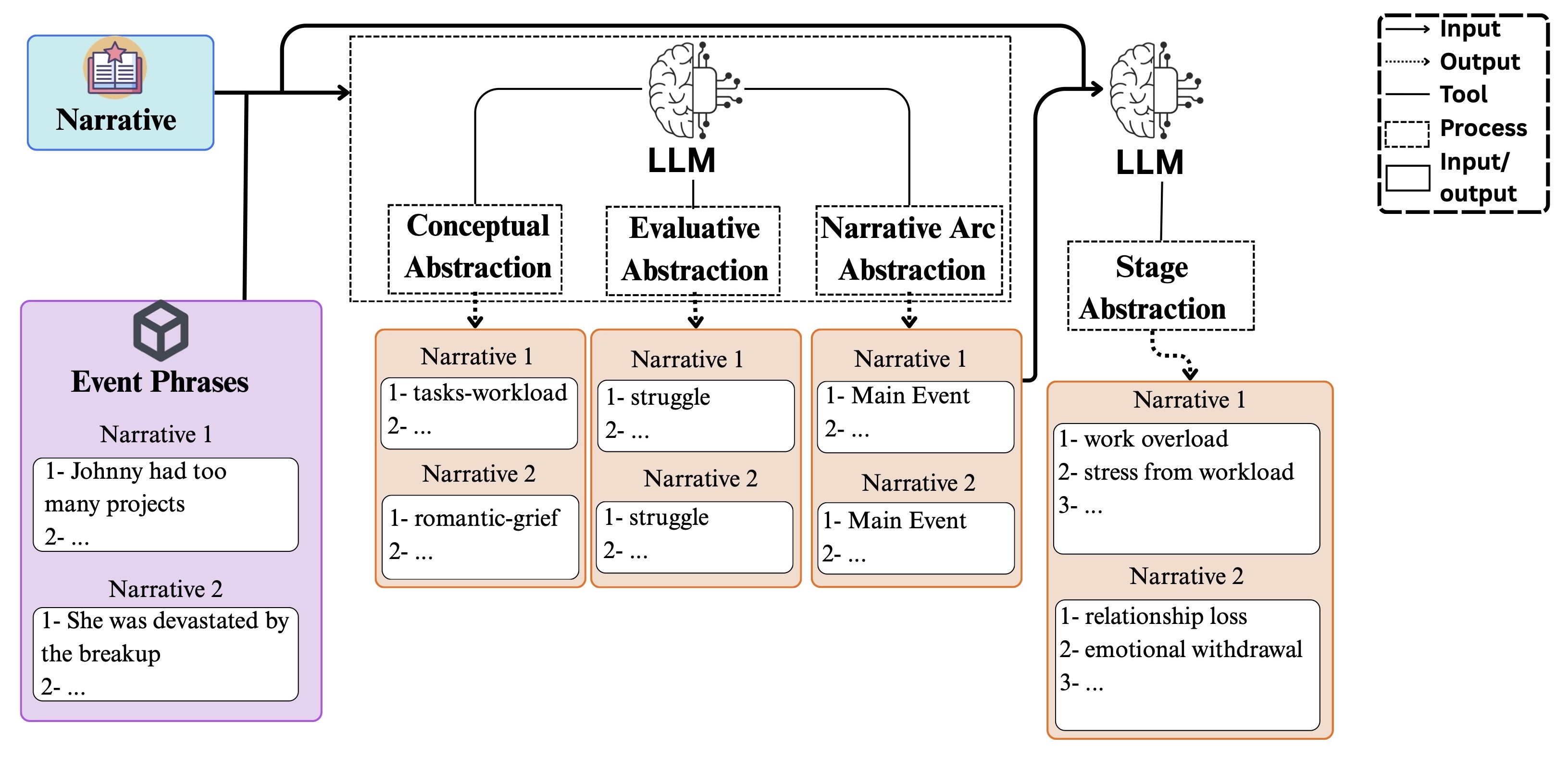} 
    \caption{
    Story Unit Abstraction. Story events are transformed into abstract representations that capture their underlying functional roles and semantic meaning. By moving beyond surface-level details, these abstractions establish the basis for structural mapping of stories.
    }
    \label{fig:framework1} 
\end{figure*}

To capture context-dependent abstractions that account for how events influence each other, we draw on framing theories \citep{sullivan2023three}, which highlight the plural nature of framing. Correspondingly, we define four abstraction levels for narrative analogy: \textbf{conceptual}, \textbf{evaluative}, \textbf{narrative arc}, and \textbf{stage} (see \autoref{tab:full-example} for examples). Formally, instead of a single abstraction set $A_s$, we introduce four sets: $A^{con}_s$, $A^{eva}_s$, $A^{arc}_s$, and $A^{stg}_s$. This design is grounded in  Sullivan's \emph{three Levels of Framing} \citep{sullivan2023three}: our conceptual abstraction aligns with their \emph{semantic} framing, our evaluative abstraction corresponds to their \emph{communicative} framing by capturing an event’s valence and functional role within the narrative, and our narrative arc and stage abstractions instantiate their \emph{cognitive} framing by capturing how events are organized and interpreted across the story. As shown in \autoref{fig:framework1}, the first three levels (conceptual, evaluative, and narrative arc) rely only on the story and extracted event phrases, whereas the stage level additionally uses the narrative arc as a previously extracted abstraction.

\noindent \textbf{Conceptual Abstraction. }Conceptual abstraction captures the semantics of an event while removing surface details such as names or domain-specific terms. We define each abstraction as $[modifier]-[root]$ (e.g., ``tasks-workload ''), where the root captures the core event meaning and the modifier provides contextual specification, similar to FrameNet frames and frame elements \citep{baker1998berkeley}. We further construct a hierarchy of conceptual abstractions with increasing levels of generality. The first level, $A^{con,0}_s$, is extracted directly from event phrases, while higher levels (e.g., $A^{con,1}_s$) are obtained by prompting the model to generalize both the event phrases and previously extracted abstractions. Formally, given a story $S$ and its event phrases $U_s = \{U_1, U_2, ..., U_n\}$, \frameworkName produces $A^{con,i}_s = \{A^{con,i}_1, A^{con,i}_2, ..., A^{con,i}_n\}$, where each event phrase maps to exactly one conceptual abstraction ($|U_s| = |A^{con,i}_s|$). Similar to event extraction, we use LLMs with few-shot prompting to extract conceptual abstraction. A summary of the conceptual abstraction prompt is provided in the appendix (\autoref{tab:cons-prompt}).

\noindent \textbf{Evaluative Abstraction. }Evaluative abstraction captures both functional roles and evaluative polarity.
Namely, we assign each event (a) a functional role: \textit{state}, \textit{action}, or \textit{outcome} \citep{arabshahi2021conversational} — and (b) an evaluative polarity—\textit{positive}, \textit{negative}, or \textit{neutral}— interpreted relative to the protagonist’s situation. An \emph{action} represents an active change, while both \emph{state} and \emph{outcome} describe conditions, with the key distinction that outcomes result from preceding events, whereas states are independent of other events. By distinguishing between independent events and those caused by others (state vs. outcome), we can potentially capture causal structure, which remains challenging and inconsistently defined in prior work \citep{wei2024llms, yang2022towards, sun2024event, hu2025large}. We assign a single joint label that combines functional roles and polarity to each event, thereby placing it into one of nine categories. 
Neutral cases follow the compositional form \emph{neutral\_state}, \emph{neutral\_action}, and \emph{neutral\_outcome}. The remaining categories are defined as follows: negative state (\emph{Struggle}), positive state (\emph{Ease}), positive action (\emph{Effort}), negative action (\emph{Indifference}), positive outcome (\emph{Gain}), and negative outcome (\emph{Loss}). Formally, this abstraction level is defined as $A^{eva}_s = \{A^{eva}_1, ..., A^{eva}_n\}$, where $|U_s| = |A^{eva}_s|$.

\noindent \textbf{Narrative Arc Abstraction. }Narrative arc abstraction captures the global organization of events and their roles in the overall story progression, beyond the local roles captured by evaluative abstraction. We define a five-stage narrative arc schema, inspired by prior work on how events unfold, interact, and lead to outcomes over the course of a story \citep{tian-etal-2024-large-language, boyd2020narrative}: a) \textit{background}: Events that provide context or prior conditions.
b) \textit{main event}: The central situation establishing the story’s core setting.
c) \textit{challenge}: The problem or obstacle faced by the protagonist.
d) \textit{action}: The protagonist’s responses or attempts to address the challenge.
e) \textit{conclusion}: The outcome or resolution of the story. Not all stories contain every stage, but the main event is mandatory, as it defines the core narrative situation. Formally, this abstraction level is defined as $A^{arc}_s = \{A^{arc}_1, ..., A^{arc}_n\}$, where $|U_s| = |A^{arc}_s|$, and each $A^{arc}_i$ is assigned to one predefined stage.

\noindent \textbf{Stage Abstraction. }Stage abstraction captures a domain-independent representation by jointly abstracting over groups of related events within each narrative stage.  We use the story, event phrases, and previously extracted abstractions as inputs and define two layers of stage abstraction. Recall that narrative arc abstraction assigns each event to one of five stages (Background, Main Event, Challenge, Action, Conclusion). In the first layer ($A^{stg,0}_s$), \frameworkName aggregates all events within the same narrative stage and extracts one abstraction that represents their shared meaning and functional role. Given a story \(S\), event phrases \(U_s = \{U_1, \ldots, U_n\}\), and the corresponding narrative arc abstraction \(A^{arc}_s = \{A^{arc}_1, \ldots, A^{arc}_n\}\), we define $A^{stg,0}_s = \{A^{stg,0}_1, A^{stg,0}_2, \ldots, A^{stg,0}_m\}$, where \(n \ge m\) and \(|U_s| \ge |A^{stg,0}_s|\). In the second layer, \frameworkName extracts a single abstraction that characterizes the narrative as a whole. This abstraction (\(A^{stg,1}_s\)), is conditioned on previously extracted layers rather than raw story text: (i) the previous level of stage abstraction \(A^{stg,0}_s\), (ii) the narrative arc abstraction \(A^{arc}_s\), and (iii) the evaluative abstraction \(A^{eva}_s\).

\begin{figure*}
    \centering 
    \includegraphics[width=\linewidth]{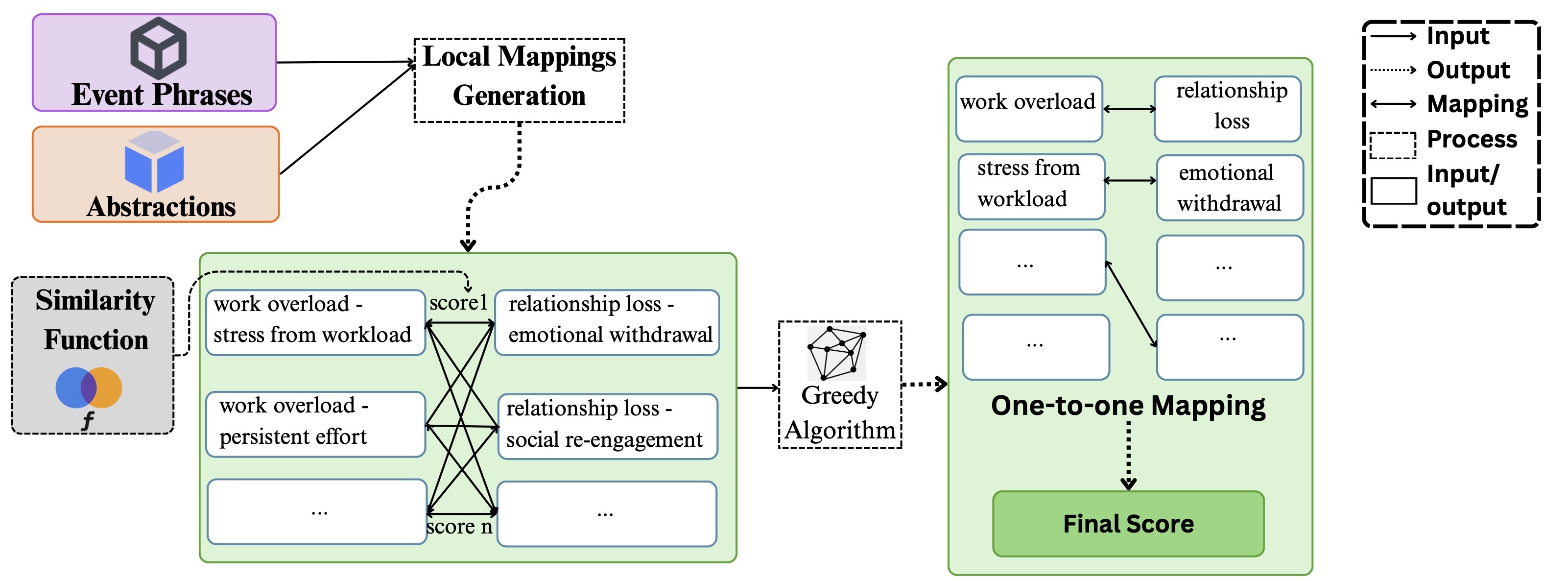} 
    \caption{Structural Mapping. For each pair of stories, all candidate mappings between units (or their abstractions) are generated and assigned similarity scores. A greedy algorithm is then used to generate a one-to-one mapping, producing a final score that reflects the overall structural correspondence between the stories.}
    \label{fig:framework2} 
\end{figure*}

\subsection{Mapping} 
\label{method:mapping}

The units and abstractions of a story $S_{i}$ form a structured representation
$R_{S_i} = \{U_{S_i}, A^{con}_{S_i}, A^{eva}_{S_i}, A^{arc}_{S_i}, A^{stg}_{S_i}\}$.
Here, $U_{S_i}$ denotes unit phrases, and $A^{con}_{S_i}$, $A^{eva}_{S_i}$, $A^{arc}_{S_i}$, and $A^{stg}_{S_i}$ denote conceptual, evaluative, narrative arc, and stage abstractions, respectively. After constructing these representations for base and candidate stories, \frameworkName performs cross-story mapping to identify the best analogical correspondence. In this paper, we deploy \textbf{structural mapping}, which generates a one-to-one mapping between units and computes a score based on mapping quality (\autoref{fig:framework2}). Similar to the FAME method extension of SME \citep{jacob-etal-2023-fame}, \frameworkName generates all possible mappings between unit pairs, assigns a score to each local mapping, and applies a greedy algorithm to find the best one-to-one mapping between units. This process produces a final score reflecting the overall mapping quality.

The first step of the mapping process is to generate pairs from each story and connect them across stories. Given a story $S_1$ and its representation $R_{S_1} = \{U_{S_1}, A^{con}_{S_1}, A^{eva}_{S_1}, A^{arc}_{S_1}, A^{stg}_{S_1}\}$, \frameworkName generates pairs using only $U_{S_1}$, $A^{con}_{S_1}$, or $A^{stg}_{S_1}$. The other abstraction levels, $A^{eva}_{S_1}$ and $A^{arc}_{S_1}$, are excluded because they consist of predefined labels that would produce repetitive or uninformative scores. Instead, they are incorporated later in the scoring function as constraints. For story $S_1$, \frameworkName generates $P_{S_1} = \{(p^1_{S_1}, p^2_{S_1}), (p^1_{S_1}, p^3_{S_1}), ..., (p^{k-1}_{S_1}, p^k_{S_1})\}$,
where each $p^i_{S_1}$ is drawn from exactly one of $U_{S_1}$, $A^{con}_{S_1}$, or $A^{stg}_{S_1}$, ensuring that pairs are generated within the same abstraction level. We repeat the same process for story $S_2$ to obtain $P_{S_2}$. Finally, \frameworkName maps all pairs from $P_{S_1}$ to all pairs from $P_{S_2}$ to form quadruples of the form
$q = [(p^i_{S_1}, p^j_{S_1}), (p^k_{S_2}, p^l_{S_2})]$, called \textbf{local mappings}.

After generating all possible quadruples, the next step is to score the quality of each local mapping. Since we expect that structural information is already captured in the abstraction layers, we use the unit pairs and their abstractions directly to compute similarity. Given a quadruple $q$ of the form $[(p^i_{S_1}, p^j_{S_1}), (p^k_{S_2}, p^l_{S_2})]$, \frameworkName computes a score that reflects how well the pair $(p^i_{S_1}, p^j_{S_1})$ from the first story corresponds to the pair $(p^k_{S_2}, p^l_{S_2})$ from the second story:

\begin{equation}
\operatorname{score}^{\text{local}}\!\left(
q
\right)
=
\frac{1}{2}\left(
\cos\!\left(\mathrm{vec}(p^i_{S_1}), \mathrm{vec}(p^k_{S_2})\right)
+
\cos\!\left(\mathrm{vec}(p^j_{S_1}), \mathrm{vec}(p^l_{S_2})\right)
\right)
\end{equation}

Recall that each $p^i$ is drawn from exactly one of the sets $U$, $A^{con}$, or $A^{stg}$. \frameworkName uses the remaining abstraction levels ($A_{eva}$ and $A_{arc}$) as \textbf{soft constraints} when computing similarity. Specifically, we retrieve the corresponding evaluative or narrative arc abstraction and append its label to each unit's text. The quadruple is therefore extended to:
$
\Big[ \big(p^i_{S_1},\, g(p^i_{S_1}),\, p^j_{S_1},\, g(p^j_{S_1})\big),\;
\big(p^k_{S_2},\, g(p^k_{S_2}),\, p^l_{S_2},\, g(p^l_{S_2})\big) \Big]$,
where $g(\cdot)$ returns the corresponding abstraction from $A_{eva}$ or $A_{arc}$. This enriched representation allows abstraction to influence similarity without strictly enforcing equality. \frameworkName then computes similarity as follows:

{\scriptsize
\begin{equation}
\begin{array}{c}
\text{score}^{\text{local}}\!\left(
q
\right)
=
\frac{1}{4}
\Big(
\cos\!\left(\mathrm{vec}(p^i_{S_1}),\, \mathrm{vec}(p^k_{S_2})\right)

+ \cos\!\left(\mathrm{vec}(g(p^i_{S_1})),\, \mathrm{vec}(g(p^k_{S_2}))\right)

+ \cos\!\left(\mathrm{vec}(p^j_{S_1}),\, \mathrm{vec}(p^l_{S_2})\right)

+ \cos\!\left(\mathrm{vec}(g(p^j_{S_1})),\, \mathrm{vec}(g(p^l_{S_2}))\right)
\Big)
\end{array}
\end{equation}
}

Finally, to construct story-level mappings, we integrate scored local mappings into a single one-to-one correspondence between units using a greedy algorithm. Each quadruple $q$, which forms a local mapping, induces two atomic mappings \(p^i_{S_1} \mapsto p^k_{S_2}\) and \(p^j_{S_1} \mapsto p^l_{S_2}\), and is associated with a score ($\text{score}^{\text{local}}(q)$), as explained before. We first sort all quadruples in descending order by their score and select the top-\(n\) as alternative starting points. For each selected quadruple \(q^{(r)}\) from the top-\(n\) alternative starting points, we initialize a partial mapping:
\begin{equation}
M^{(r)}_0 = \{\, p^i_{S_1} \mapsto p^k_{S_2},\; p^j_{S_1} \mapsto p^l_{S_2} \,\}.
\end{equation}
\frameworkName then iterates over the remaining quadruples in descending score order and greedily add a quadruple \(q\) to the current mapping \(M^{(r)}\) if it is compatible with the existing one-to-one constraint: for each induced mapping \(x \mapsto y\) from \(q\), either \(x\) is not yet mapped in \(M^{(r)}\) and \(y\) is not assigned elsewhere, or \(x\) is already mapped to the same \(y\). When a compatible quadruple is added, its induced correspondences are merged into \(M^{(r)}\). This process continues until no further quadruples can be added without violating the one-to-one constraint. For each starting quadruple \(q^{(r)}\), we obtain a \textbf{global mapping} with an associated score:
\begin{equation}
\mathrm{Score}^{(r)}(S_1,S_2) = \sum_{q \in \mathcal{Q}^{(r)}} s(q),
\end{equation}
where \(\mathcal{Q}^{(r)}\) is the set of quadruples included in the mapping. The final story-level score is defined as the maximum score across all \(n\) runs,
\begin{equation}
\mathrm{Score}(S_1,S_2) = \max_{r \in \{1,\dots,n\}} \mathrm{Score}^{(r)}(S_1,S_2).
\end{equation}

\frameworkName differs from FAME in two key aspects. First, the representation of unit pairs: FAME operates on entities, whereas \frameworkName uses event phrases or their abstractions as units. Second, the scoring mechanism of mappings: FAME extracts relations between entities with an LLM and scores mappings through clustering over these relations, while \frameworkName scores mappings directly on event phrases and abstractions, embedding structural information within the abstractions themselves. We also introduce the option to incorporate additional information as soft constraints, which is not present in FAME. Meanwhile, the global mapping procedure remains the same, as we adopt the same greedy algorithm used in the FAME framework.

\section{Experimental Setup}
\label{sec:setup}

\paragraph{Implementation details} We instantiate \frameworkName by using either Qwen3-8B \citep{yang2025qwen3} and Llama-3.1-8B \citep{grattafiori2024llama}, both recent and competitive open-source LLMs widely used in evaluation studies, to extract event phrases and abstractions from the stories. To \textit{extract event phrases}, the model is instructed to decompose each story into atomic events representing actions, states, observations, intentions, decisions, or reasons. Each event is expressed as a short, standalone phrase that preserves narrative order, ensures coverage, and avoids redundancy. Because textual order may differ from chronological order, the model is also asked in a different prompt to assign each event a temporal index starting from 1 that reflects its actual sequence, with consecutive indices for simultaneous events. For \textit{conceptual} abstraction, we prompt the model to assign each event a semantic frame of the form $[modifier]-[root]$, where the root captures the core event or state and the modifier provides a single-word contextual qualifier. We use LLM prompting instead of existing semantic parsers \citep{das2014frame, palmer2005proposition, banarescu2013abstract} because they cannot produce context-dependent narrative abstractions. We also adopt an open-vocabulary approach, rather than relying on closed resources like FrameNet, due to their limited coverage of narrative events and the unreliable frame inference by LLMs \citep{guo2024nutframe}, as confirmed by our preliminary experiments. For \textit{evaluative and narrative arc} abstractions, we provide definitions of our categories and ask the model to assign each event to one of the predefined roles or stages. Finally, for \textit{stage} abstraction, event phrases are grouped by stage and the model generates a single abstract label per stage that generalizes beyond the member phrases while avoiding story-specific details.

All model inference is performed with the vLLM library \citep{kwon2023efficient} for efficient and scalable deployment. Our prompts include example stories from the ROCStories dataset \citep{mostafazadeh2016corpus} as few-shot demonstrations, adapted to match the style and structure of our datasets. Except for event extraction, all tasks use CoT prompting \citep{wei2022chain}, where the model first provides a brief rationale in a specified format and then outputs the final answer. All final outputs are returned in JSON format for easy storage and processing. We use the all-MiniLM-L6-v2 model \citep{wang2020minilm} from Sentence-Transformers \citep{reimers2019sentence} to compute unit embeddings, which are compared using cosine similarity. All experiments were conducted on NVIDIA A100 GPUs on the DAS-6 \citep{bal2016medium} and Snellius high-performance computing clusters.\footnote{https://servicedesk.surf.nl/wiki/}

\paragraph{Hyperparameters} For all LLM-based experiments, we used greedy decoding with a temperature of 0 to ensure deterministic behavior. In the greedy algorithm to generate the global mapping, we set the output size to $n = 3$ for the ARN dataset and $n = 2$ for the MCQ dataset, retaining the top-3 global mappings for ARN and the top-2 for MCQ. These values were chosen through hyperparameter tuning to balance performance and computational cost. We use a larger beam size for ARN due to its longer and more complex narratives.

\paragraph{Evaluation protocol} Given a base story and several candidate target stories, we aim to identify the target that produces the best structural mapping with the base story. For each base story, we ask \frameworkName to select one target story as its prediction. We evaluate each approach using \textit{accuracy}, calculated by comparing the predicted targets with the ground-truth targets across all examples.

\begin{table}[t]
\caption{Statistics of the MCQ and ARN datasets.}
\label{tab:dataset-overview}
\footnotesize
\centering
\setlength{\tabcolsep}{4pt}
\renewcommand{\arraystretch}{0.95}

\begin{tabular*}{\linewidth}{@{\extracolsep{\fill}} l r r r}
\hline
\textbf{Dataset name} & \textbf{\#Target stories} & \textbf{Dataset size} & \textbf{Avg. length (\#tokens)} \\
\hline
MCQ & 4 & 360 & 17.3 \\
ARN & 2 & 1,096 & 65.8 \\
\hline
\end{tabular*}

\end{table}

\paragraph{Benchmarks} For our experiments, we used two recently introduced benchmarks for analogical reasoning: (a) StoryAnalogy-MCQ (MCQ), which reformulates the 24K Story Analogy \citep{jiayang-etal-2023-storyanalogy} dataset into 360 multiple-choice questions with carefully designed distractors, and (b) Analogical Reasoning on Narratives (ARN), which contains 1,096 narrative pairs based on proverbs \citep{sourati-etal-2024-arn}. Table \ref{tab:dataset-overview} shows statistics of the two datasets. The MCQ dataset provides four candidate targets for each base story, while the ARN dataset provides two. The final column shows that the ARN stories are about four times longer on average than the MCQ stories.

A key property of the ARN dataset is its four partitions. For each base story, there are two candidate targets: the correct answer (analogy) and the incorrect option (disanalogy). Both can either share surface-level similarity with the base story (near) or lack it (far). Combining these two dimensions yields four categories (the first for the analogy and the second for the disanalogy): near-far, far-far, near-near, and far-near. The near-far category is the easiest, as the correct analogy is structurally and superficially similar, while the disanalogy is clearly different. In contrast, far-near is the most challenging: the analogy lacks surface similarity, and the disanalogy shares surface level with the base story. For example, \autoref{fig:example} illustrates a far-near category: the base story and the disanalogy ($S_{T1}$) share surface similarity, both involving a workspace problem, whereas the base story and the analogy ($S_{T2}$) differ on the surface, with one about a workspace issue and the other about a breakup.

\paragraph{Baselines} We compare \frameworkName against baseline models that receive the extracted events and select the best target story instead of computing a numerical score. We evaluate the same two LLMs: Qwen3-8B and Llama-3.1-8B. We consider the following instruction style: the model is asked to perform event-level mapping by comparing events across stories based on their abstract meaning, functional role, and narrative position, and then to select the target with the strongest structural alignment. We evaluate this using zero-shot and zero-shot chain-of-thought (CoT) prompting \citep{wei2022chain}.

\section{Results}
\label{results}

In this section, we present the results of \frameworkName with an emphasis on our four abstraction levels. Additionally, we demonstrate the inconsistency in prompt-based LLM analogical reasoning across different prompts and input units, and conclude with an error analysis that highlights remaining challenges. The systematic experiments is this section are made possible by the modular design of our \frameworkName pipeline.

\begin{table*}[t]
\caption{\textbf{Structural mapping with abstraction provides competitive results compared to LLM analogical reasoning.} This table reports the results of LLM when prompted to perform event mapping with zero-shot and chain-of-thought settings, structural mapping over event phrases without abstraction, and structural mapping with different abstraction combinations. While structural mapping without abstraction~(\textcolor{gray}{gray rows}) performs poorly and even falls below random, adding abstraction substantially improves performance, yielding results competitive with LLM event mapping. The input for all experiments is \emph{event phrases}, and \textit{SM} stands for Structural Mapping. An empty column indicates that the corresponding abstraction was not used for that row. The highest score is \textbf{bolded}, and the second-highest is \underline{underlined} for each model and dataset split.}
\label{tab:main_table}
\centering
\normalsize
\setlength{\tabcolsep}{10pt}
\newcommand{\hdr}[2]{
  \makecell[c]{\rule{0pt}{2.8ex}\textbf{#1}\\\textbf{#2}\rule[-1.2ex]{0pt}{0pt}}
}
\newcommand{\hdright}[2]{
  \makecell[r]{\rule{0pt}{2.8ex}\textbf{#1}\\\textbf{#2}\rule[-1.2ex]{0pt}{0pt}}
}
\newcommand{\ccg}{\cellcolor{lightgrayrow}}
\newcommand{\ccb}{\cellcolor{lightbluerow}}

\renewcommand{\arraystretch}{1.0}
\setlength{\extrarowheight}{1pt}

\makebox[\textwidth][c]{
\resizebox{0.95\textwidth}{!}{
\begin{tabular}{c c c c c c r r r}
\hline
 & \multicolumn{4}{c}{\textbf{Abstraction}} &  &  &  \\
\cline{2-5}
 & \textbf{Conceptual} & \textbf{Evaluative} & \textbf{Stage} & \textbf{Narrative Arc}
 & \textbf{Mapping} & \textbf{MCQ} &  
 \multicolumn{2}{c}{\textbf{ARN}} \\
\cline{8-9}
 &  &  &  &  & & &
 \hdright{Near}{Analogy} & \hdright{Far}{Analogy} \\
\hline

% ===================== Qwen3-8B =====================

\multirow{6}{*}{\rotatebox[origin=c]{90}{\textbf{Qwen3-8B}}}
&  &  &  &  & LLM-ZS &  0.37 & 0.52 & \textbf{0.52}  \\
&  &  &  &  & LLM-CoT &  \underline{0.41} & \textbf{0.79} & 0.42  \\
\cdashline{2-9}
  & \ccg - & \ccg - & \ccg - & \ccg - & \ccg SM & \ccg 0.17 & \ccg 0.67 & \ccg 0.24 \\
& $\checkmark$ &  &  &  & SM & \underline{0.41} & 0.67 & 0.46  \\
& \ccb $\checkmark$ & \ccb $\checkmark$ &  \ccb & \ccb & \ccb SM & \ccb \textbf{0.46} & \ccb 0.63 & \ccb \underline{0.47} \\
&  &  & $\checkmark$ & & SM & 0.29 & \underline{0.69} & 0.40 \\
&  &  & $\checkmark$ & $\checkmark$ & SM & 0.30 & 0.62 & 0.45 \\
\cline{1-9}

% ===================== Llama3.1-8B =====================

\multirow{6}{*}{\rotatebox[origin=c]{90}{\textbf{Llama3.1-8B}}}
&  &  &  &  & LLM-ZS & 0.23 & \textbf{0.76} & 0.42 \\
&  &  &  &  & LLM-CoT & 0.28 & \underline{0.75} & 0.41 \\
\cdashline{2-9}
 & \ccg - & \ccg - & \ccg - & \ccg - & \ccg SM & \ccg 0.16 & \ccg 0.66 & \ccg 0.20 \\
& $\checkmark$ &  &  &  & SM & \underline{0.42} & 0.64 & \textbf{0.46}  \\
& \ccb $\checkmark$ & \ccb $\checkmark$ & \ccb & \ccb &  \ccb SM & \ccb \textbf{0.45} &  \ccb 0.62 & \ccb \underline{0.45} \\
&  &  & $\checkmark$ & & SM & 0.24 & 0.73 & 0.36 \\
&  &  & $\checkmark$ & $\checkmark$ & SM & 0.27 & 0.65 & 0.43  \\
\cline{1-9}

\end{tabular}
}%
}
\end{table*}

\subsection{Main Results}
\label{sec:main-result}

\noindent \textbf{How well does our \frameworkName structural mapping pipeline with different abstraction levels perform on narrative analogy compared to prompt-only LLMs?}
\autoref{tab:main_table} compares the accuracy of analogical reasoning by LLMs alone and when enriched by structural mapping (SM), with or without abstraction. SM without abstraction performs very poorly, even below random. For Qwen, accuracy on MCQ is 0.17, below the 0.25 random baseline, and 0.24 on ARN far analogy, also below the 0.50 random baseline. As structural mapping compares event phrases using cosine similarity over text-based embeddings, it suffers from the following limitation: distributional embeddings mainly capture surface-level semantics and often miss deeper aspects such as functional roles, causal structure, or logical relations like negation and contradiction \citep{aina2018distributional, nikolaev2023universe, steck2024cosine}. Consequently, adding abstractions substantially improves performance. On MCQ, our best result reaches 0.46 with Qwen, exceeding the best LLM-only result of 0.41 by 0.05. For Llama, \frameworkName obtains 0.45, compared to 0.28 with LLM-only mapping (+0.17). These results show that structural mapping with abstraction outperforms LLM analogical reasoning on MCQ for both models. On ARN, the pattern is more nuanced. For Qwen, \frameworkName achieves the second-best results in both near and far categories: it is behind CoT on near analogies (0.69 vs.\ 0.79) and zero-shot on far analogies (0.47 vs.\ 0.52). If we compare specifically to CoT, \frameworkName performs worse on near analogies but better on far ones. For Llama, zero-shot performs best on near analogies (0.73 vs 0.76), while \frameworkName outperforms on far analogies (0.46 vs.\ 0.42). This gap in the ARN dataset reflects different strengths: \emph{LLMs perform better on near analogies, while \frameworkName performs better on far (more difficult) analogies.}

\noindent \textbf{What is the impact of different abstraction levels on performance?}
Since replacing event phrases with abstractions consistently improves performance, we next compare the impact of different abstractions. On MCQ, the best results are achieved using conceptual abstraction with evaluative abstraction as a constraint: for Qwen, the accuracy increases from 0.17 to 0.46, and for Llama from 0.16 to 0.45 (+0.29). Conceptual abstraction alone also yields competitive results, but lower than when evaluative constraints are added (e.g., for Qwen, 0.41 without evaluative vs.\ 0.46 with evaluative). This suggests that MCQ benefits from conceptual representations, and evaluative abstraction adds functional signals that further guide mapping. On ARN, however, the best results come from using conceptual abstraction alone, while adding evaluative constraints slightly reduces performance (e.g., for Qwen, 0.67 without evaluative on near analogies vs.\ 0.63 with evaluative). This is unexpected, since many ARN examples involve shifts in event valence, where state/action/outcome roles and polarity should help alignment. We discuss the reason for this in more detail in \autoref{sec:error_analysis} and \autoref{sec:discussion}.

The results with stage abstraction and narrative arc abstraction show an interesting pattern. Stage abstraction groups multiple events by their role in the story, and we map these stages using structural mapping. We also apply narrative arc labels as soft constraints to restrict which stages can align. On ARN, performance is close to conceptual abstraction (e.g., for Llama: 0.65 with stage abstraction, with arc constraints, compared to 0.64 with conceptual abstraction without evaluative, and 0.62 with evaluative on near analogies). On MCQ, however, performance drops substantially (e.g., for Llama: from 0.45 with conceptual abstraction to 0.24 and 0.27 with stage abstraction). This aligns with our expectations. MCQ stories are short and lack clear narrative arcs, so stage abstraction offers little benefit compared to using only event phrases. On ARN, where arcs are more developed, stage abstraction improves over raw event phrases but does not surpass conceptual abstraction. This may reflect limitations in the stage abstraction itself, the mapping strategy, or both, which we examine further in \autoref{sec:error_analysis} and \autoref{sec:discussion}.

\subsection{Ablation Studies}
\label{sec:ablation}

\begin{figure*}
    \centering 
    \includegraphics[width=0.9\linewidth]{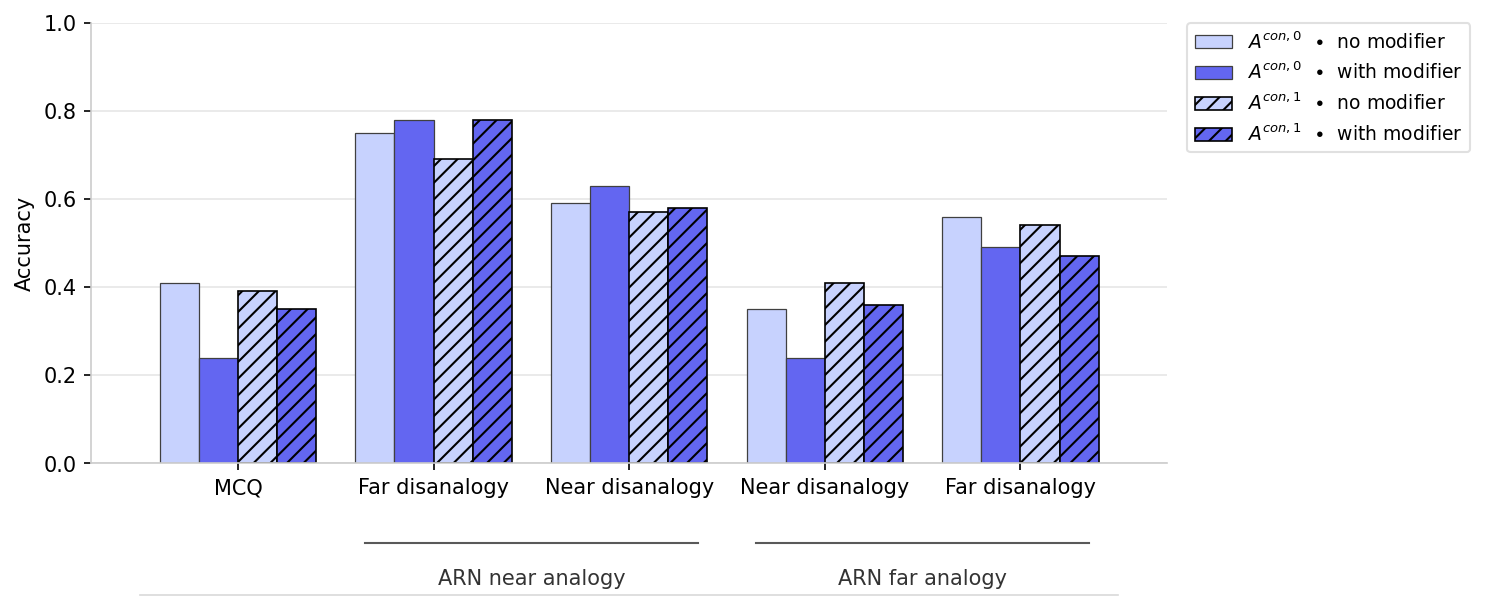} 
    \caption{\textbf{Different levels of conceptual abstraction affect performance depending on the degree of surface similarity.} This figure shows the effect of hierarchical conceptual abstraction levels for Qwen, with and without modifiers. Solid bars denote $A^{con, 0}$, hatched bars $A^{con, 1}$; pale indigo bars use only the root, and indigo bars use both the modifier and the root. The first pale indigo solid bar in each group corresponds to the setting in \autoref{tab:main_table}.}
    \label{fig:cons-setting-qwen} 
\end{figure*}

\noindent \textbf{How do different settings of conceptual abstraction affect performance?}
In \autoref{method:abstraction}, we describe conceptual abstraction as extracting a modifier and a root for each event, with two hierarchical levels ($A^{con, 0}$ and $A^{con, 1}$). In \autoref{tab:main_table}, we used only the root of $A^{con, 0}$. A natural question is how different settings of this abstraction affect performance. Results for Qwen are shown in \autoref{fig:cons-setting-qwen} (Llama results appear in the appendix, \autoref{fig:cons-setting-llama}). On MCQ, the best result comes from using only the root of $A^{con, 0}$ (0.41 for Qwen), suggesting two points: 1) modifiers are unnecessary for this dataset, and 2) further generalizing the root from $A^{con,0}$ to $A^{con,1}$ does not improve performance. For ARN near analogies, the best results come from $A^{con, 0}$ with both modifier and root (e.g., near-far: 0.78 for Qwen, 0.75 for Llama), indicating that retaining domain-specific information helps. For far analogies, usually the best results come from using only the root without any modifiers, indicating that modifiers are not helpful for this type of analogy. For far-near cases, which are the most difficult category, the best results come from $A^{con, 1}$ using only the root (0.41 for Qwen, 0.42 for Llama). These cases require minimizing surface cues; higher abstraction and removing modifiers are most effective. However, performance remains below the 50\% random baseline across all settings. This suggests that either the abstraction hierarchy is still insufficient or that structural mapping with embedding and cosine similarity alone cannot handle far-near cases. We discuss this further in \autoref{sec:error_analysis}. For far-far cases, results are mixed: Qwen performs best with roots from $A^{con, 0}$ (0.56), while Llama benefits from the more generalized $A^{con, 1}$ with a modifier and root (0.58). Since neither the correct answer nor the distractors show strong surface similarity, and modifiers introduce domain-specific information, using the modifiers may or may not help.

\begin{figure*}
    \centering 
    \includegraphics[width=0.8\linewidth]{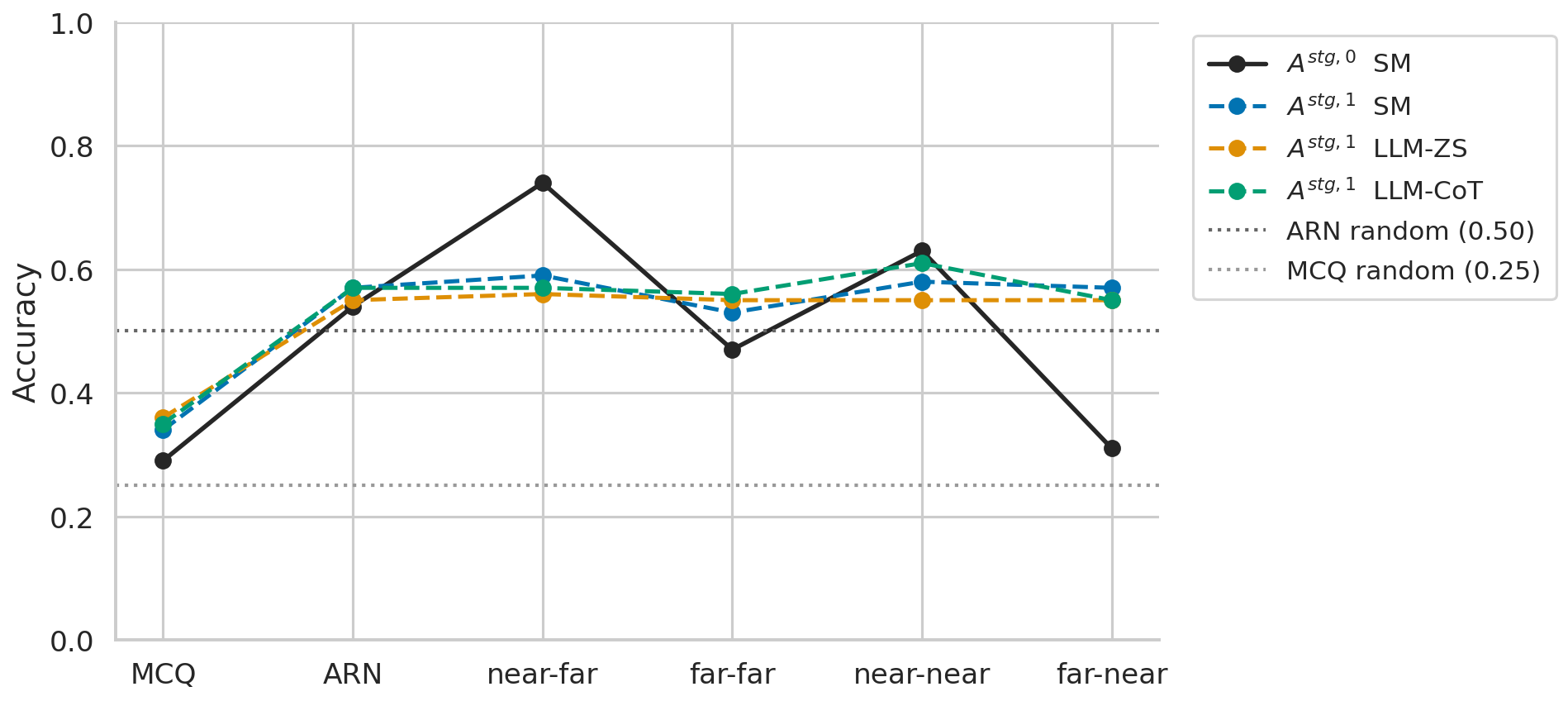} 
    \caption{\textbf{Stage abstraction effectiveness depends on the type of analogy, with lower levels benefiting near cases and higher levels improving far cases.} Effect of hierarchical stage abstraction levels with different settings for Qwen. The solid line shows $A^{stg,0}$ (also reported in \autoref{tab:main_table}), while dashed lines show $A^{stg,1}$ under three mapping settings: SM (which reduces to cosine similarity since each story is a single phrase), and LLM zero-shot and CoT, where phrases are treated as high-level messages following \citep{sourati-etal-2024-arn}. Horizontal lines mark random performance for MCQ and ARN. }
    \label{fig:stg-setting-qwen} 
\end{figure*}

\noindent \textbf{How do stage abstraction levels affect performance?}
Similar to conceptual abstraction, stage abstraction also has two hierarchical levels ($A^{stg,0}$ and $A^{stg,1}$). We analyze their impact in \autoref{fig:stg-setting-qwen} (Llama results appear in \autoref{fig:stg-setting-llama}). On MCQ and the aggregated ARN results, differences are small. However, category-level results on ARN reveal clearer trends. For near analogies (near-far and near-near cases), $A^{stg,0}$ performs substantially better. In contrast, for far analogies (far-far and far-near), $A^{stg,1}$ achieves the strongest results. Notably, this is the first setting in which the performance of \frameworkName exceeds random across all ARN categories (for Qwen in all mapping settings and for Llama with LLM-CoT). These results show that $A^{stg,1}$ reduces differences across categories, bringing near and far cases closer together. Additionally, the results indicate that the effectiveness of stage abstraction depends on the type of analogy. Lower-level abstraction ($A^{stg,0}$) benefits near analogies by preserving some surface detail, while higher-level abstraction ($A^{stg,1}$) improves performance on far analogies by suppressing surface cues and aligning stories at a more general level, consistent with the construction of the ARN dataset around shared proverbs \citep{sourati-etal-2024-arn}.

\begin{figure*}
    \centering 
    \includegraphics[width=0.9\linewidth]{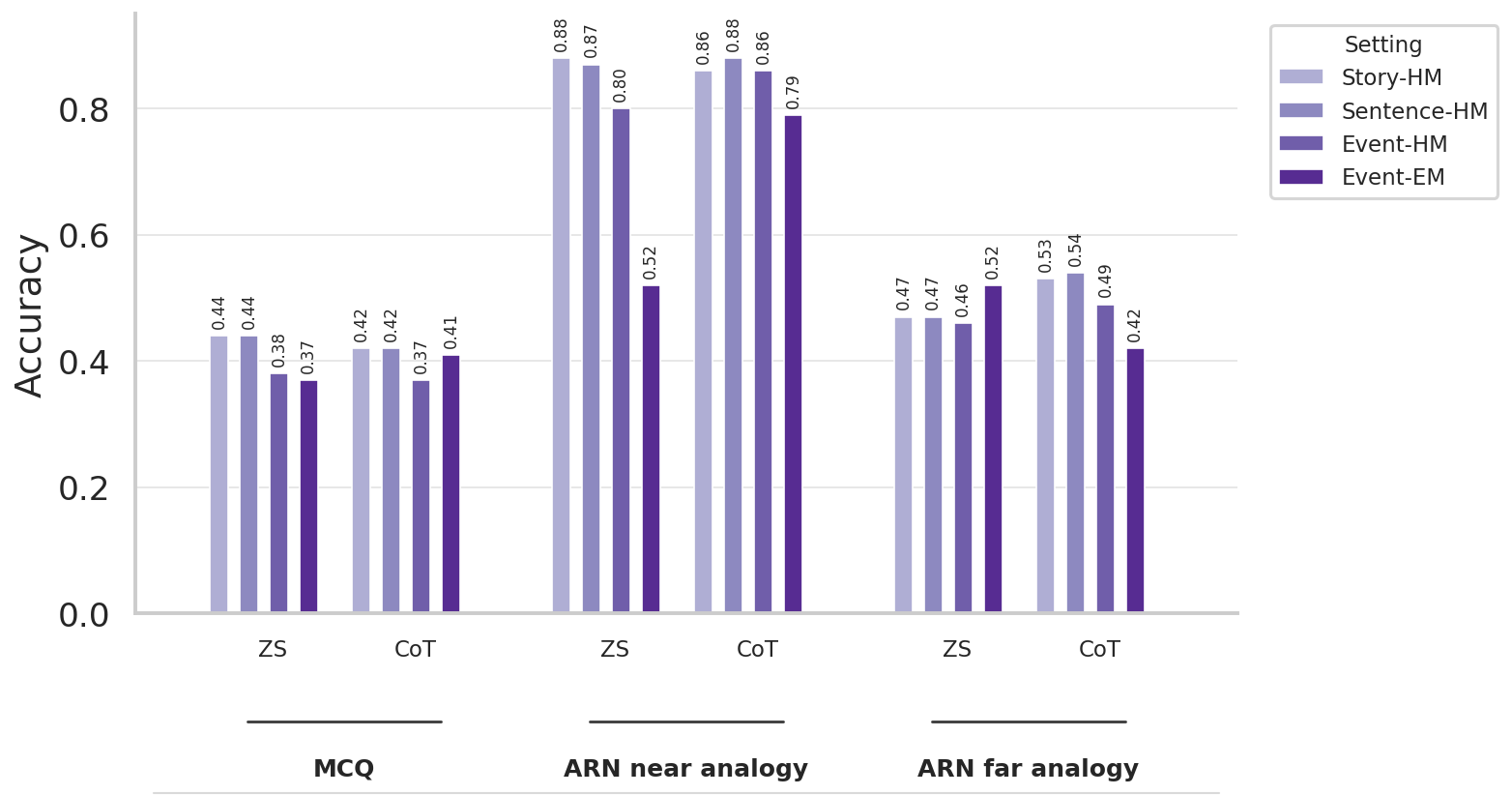} 
    \caption{\textbf{LLMs’ analogical reasoning is inconsistent across different input units and instructions.} The result of using different input and mapping instructions with Qwen for analogical reasoning.}
    \label{fig:consistent-qwen} 
\end{figure*}

\noindent \textbf{Are LLMs consistent and reliable in performing analogical reasoning?}
\frameworkName treats event phrases as story units over which LLMs perform mapping. 
However, prior work, such as the ARN dataset paper \citep{sourati-etal-2024-arn} feeds full stories to the model and treats high-level message inference as a proxy for analogical reasoning. This raises two questions: \textit{does the granularity of input units matter}, and \textit{are LLMs consistent across different unit granularities?} Here, we compare three inputs: full stories, sentences, and event phrases, under different prompts (results for Qwen in \autoref{fig:consistent-qwen}, Llama in \autoref{fig:consistent-llama}). For stories and sentences, we follow ARN's setup and ask for high-level message inference; for event phrases, we test both high-level message inference and event mapping. Performance with stories and sentences is very similar, as expected, since sentence splitting mainly changes formatting (e.g., Qwen with CoT on ARN near: 0.88 vs. 0.87). Still, differences of 2–5\% appear in some cases (e.g., Llama with CoT on MCQ: 0.39 vs. 0.34). When switching to event phrases, performance is more affected and often drops (e.g., Qwen + ZS on MCQ: 0.44 vs. 0.38), though occasional improvements occur (e.g., Llama zero-shot on ARN far: 0.50 vs. 0.52). These mixed patterns suggest that while switching to event phrases generally does not remove crucial information, it still alters the model’s reasoning. Qualitative analysis reveals a curious pattern: models infer similar high-level messages across unit representations but arrive at different answers (see \autoref{tab:qualitive-example} in the appendix for an example). This indicates sensitivity to input format rather than consistent reasoning, aligning with prior findings on the lack of robustness of LLMs \citep{errica2025did}. 

We then compare high-level message inference with asking for event mapping. Across most settings, switching from message inference to event mapping causes a sharp drop (e.g., Qwen zero-shot on ARN near with sentences: 0.88 vs. 0.52), though some improvement can also be seen (e.g., Qwen zero-shot on ARN far with sentences: 0.47 vs. 0.52). These results indicate that high-level message inference is not equivalent to analogical reasoning. While models may infer messages using learned shortcuts, requiring explicit mapping exposes weaknesses in structural alignment. We explore this nuance further in the next section.

\subsection{Error Analysis}
\label{sec:error_analysis}

In the following, we present a detailed error analysis based on our experimental observations. We group the errors into four categories: event extraction, abstraction, mapping, and benchmark factors. As in \autoref{sec:ablation}, this detailed analysis of error sources is only possible because of the structured and modular nature of our \frameworkName pipeline.

\noindent \textbf{Event extraction errors.}
While LLMs can generally extract event phrases effectively, we observe four recurring issues: (i) \textit{Missing events}: Some events are omitted entirely, e.g., from ``Jucy should be content of what Maria gives to her at least Maria is thinking of her,'' the model extracted only ``Maria thinks of Jucy,'' missing ``Jucy should be content.'' Although treating the phrase ``Jucy should be content'' as an event may be debatable, we consider it an event based on our definition in \autoref{method:unit-extraction} and \autoref{sec:setup}. (ii) \textit{Altered meaning:} The extracted event sometimes changes key details, such as generating ``Worker had just graduated from school'' instead of ``Neighbor had just graduated from school.'' (iii) \textit{Redundant events:} The same event may appear in slightly different forms because the model duplicates it, e.g., ``Flower wilted faster than others'' and ``Flower wilted faster than morning pollinated flowers,'' which convey essentially the same meaning. (iv) \textit{Hallucinated events:} The model may introduce events not stated in the story. For instance, from ``Husband realized that her wife is better and a good wife to be with,'' it generated ``Husband decided to be with his wife,'' adding an unstated decision. Although infrequent, these errors demonstrate that even basic event extraction is non-trivial and can introduce noise into the pipeline, which the remaining components must account for. Similar limitations are highlighted in prior work on extracting event triggers, types, and arguments \citep{li-etal-2023-open}.

\noindent \textbf{Event abstraction errors.}
Event abstraction is a novel task and is arguably much harder than extraction, as it requires LLMs to understand how events relate, what their general meaning is, and what role they play in the narrative.  Our analysis shows that LLMs struggle with abstraction, particularly when causality is central. We observe three main issues: (i) \textit{Staying close to the surface:} conceptual abstractions are sometimes not sufficiently generalized. For example, from ``Their second anniversary is tomorrow,'' the model produced ``relationship-anniversary.'' While not wrong, it largely repeats the surface term and fails to capture a higher-level meaning. (ii) \textit{Missing the pragmatic role of an event:} in other cases, abstractions are general but fail to capture the core function of the event within the story. For example, from ``Lily told a joke,'' the model generated ``message speech.'' While this is a fair abstraction, it misses the intent in the given story, namely, using humor to overcome difficulty or sustain engagement. (iii) \textit{Misinterpreting causal cues}: evaluative and narrative arc abstractions require identifying whether an event is a state or an outcome and whether it is positive or negative. For instance, ``my jacket tore apart'' can be a state or an outcome depending on prior causes. Similarly, determining polarity or narrative role also depends on the causal context. Despite careful prompting, models frequently misclassify such roles, revealing persistent weaknesses in causal reasoning. This aligns with recent findings that LLMs struggle to apply and transfer causal relations in analogical reasoning \citep{lee2025curious}.

\noindent \textbf{Mapping errors.} While structural mapping is an established method for analogical reasoning and holds a promise for structural analogies in stories, our investigation reveals two main limitations: (i) \textit{Brittleness of the similarity function:} As discussed in \autoref{sec:main-result}, in our work, we use the FAME framework as an implementation of structural mapping, which relies on distributional embeddings, which mainly capture surface patterns \citep{aina2018distributional, nikolaev2023universe, steck2024cosine}. As a result, two dissimilar phrases may receive a high similarity score simply because they share related words (e.g., ``reward for hard work'' vs. ``rejection after hard work''). (ii) \textit{Inability to model different patterns of story analogies:} Multiple valid structural patterns can lead to the correct answer. In ARN, two stories may be analogous even when corresponding events have opposite roles or outcomes. For example, see the third pattern in \autoref{tab:patterns}, where the two analogous stories have opposite outcomes: one involves having unfaithful children, while the other involves having faithful children. SM's greedy global mapping enforces a single one-to-one alignment and cannot flexibly capture such variations.

\begin{figure*}
    \centering 
    \includegraphics[width=0.9\linewidth]{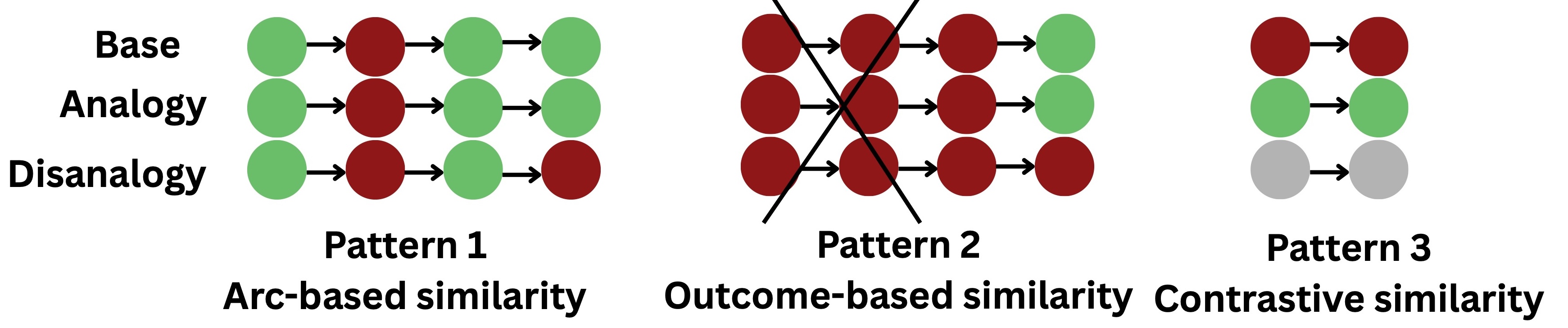} 
    \caption{Different patterns appear in the ARN stories when identifying the correct answer. Pattern 1 emphasizes the journey itself, meaning that the order and progression of events are crucial. Pattern 2 focuses mainly on the final outcome, where the sequence of events plays a minor role. Pattern 3 highlights the same underlying rule by presenting opposite or contrasting sequences, where similarity emerges through contrast rather than surface alignment.}
    \label{fig:patterns} 
\end{figure*}

\noindent \textbf{Benchmark factors. }
We identified several patterns in the employed benchmarks that raise doubts about their suitability for assessing analogical reasoning. On the one hand, in MCQ, the correct answer can often be selected based on grammatical similarity rather than structural reasoning. Consider the base story: ``The judge made a sentence for the crime. The convicted person was to face the consequences.'' The analogous story: ``The doctor made a diagnosis of the disease. The treatment plan was set in motion,'' closely mirrors the grammatical structure of the base story. Although this is a far analogy, where both stories involve addressing a problem (crime or disease) followed by its consequences, a model or a human could pick the answer by matching patterns rather than structural mapping.

On the other hand, the longer and richer stories in ARN introduce a different challenge. ARN analogies are based on proverbs, and analogical ability is defined as mapping narratives based on the high-level message they convey. In this view, analogical mapping is equated with matching high-level messages related to shared moral or common knowledge themes. We argue that this view is problematic: while a high-level message may be inferred from analogical reasoning, analogical reasoning typically involves aligning structured relations and event sequences, not just matching final interpretations. Our further analysis reveals multiple narrative arc patterns that enable identification of the correct answer in ARN (illustrated in \autoref{fig:patterns}, and with examples in \autoref{tab:patterns}). The first pattern aligns with a structural view of analogy, where the sequence of events is central. However, the other two patterns deviate from this view. In Pattern 2, only the final outcome matters, and the preceding sequence is largely irrelevant. In Pattern 3, the same underlying rule is expressed through opposite sequences and outcomes. The latter two patterns cannot be captured by standard structural mapping; they require reasoning about contradiction and opposition, for example, through approaches based on Natural Language Inference (NLI) \citep{williams2018broad}.

\section{Discussion and Future Directions}
\label{sec:discussion}

In this paper, we propose the first model that enhances structural mapping with LLMs to perform analogical reasoning in narratives. Section \ref{results} shows that while LLMs perform well on near analogies, they struggle when structural mapping is required, and surface similarity is weak. Our \frameworkName framework, which integrates structural mapping across multiple levels of abstraction, performs worse than LLMs on near analogies but improves performance on far analogies. We also examine how different abstraction levels affect performance and show that certain settings can achieve above-random performance across challenging categories. Finally, we present a detailed error analysis that highlights limitations beyond the scope of this work, including LLM errors in extraction and abstraction, challenges in the mapping component during similarity computation, and differences in analogy patterns across datasets.

Next, we reflect on the design components of our \frameworkName framework, highlighting its limitations, challenges, and directions for future work. We identify four main questions that stem from the observations in the previous section.

\noindent \textbf{What is the optimal formalism to represent story units?}
In \autoref{method:unit-extraction}, we outlined the rationale for decomposing stories into units, the constraints governing this decomposition, and our motivation for selecting event phrases as the unit representation. In the spirit of the predicate-argument representations in SME, future work should consider alternative representations that may capture story structure while remaining amenable to reliable and effective mapping. One promising alternative is Abstract Meaning Representation (AMR). AMR graphs have recently demonstrated improvements across diverse tasks, including question answering, machine translation, and information extraction \citep{wein2024survey}, by providing explicit, structured encodings of sentences' semantic content \citep{lee2021maximum}. Answer Set Programming (ASP) presents another compelling direction, offering declarative semantics well-suited to structured reasoning; recent work has explored using LLMs to translate natural language into ASP formalisms for tasks such as story generation and question answering \citep{wang2024guiding, santana2025question}. A natural extension of our work would be to convert entire stories into such structured representations and perform analogical reasoning directly over them to identify the correct target. Integrating structured representations would likely require different graph matching methods \citep{jiang2025comprehensive, zhu2024survey}, potentially extending or replacing our current structural mapping.

In preliminary experiments, we explored both AMR and ASP representations. For AMR, we converted stories into graphs and computed similarity via graph edit distance \citep{jain2024graph}, yielding strong results on MCQ (approximately 0.6, versus 0.46 in our current pipeline), suggesting that MCQ relies heavily on grammatical similarity, but lower performance on ARN, where capturing high-level narrative meaning proved difficult when deriving abstractions from graph structures. ASP representations appeared more promising initially, as models produced well-formed logical representations, yet defining abstractions that encode higher-level meaning within this formalism remains elusive. 

\noindent \textbf{How can we effectively derive and reason with abstractions in narratives?}
\autoref{results} confirmed that decomposing a story into textual units is not sufficient on its own to perform analogical reasoning and requires abstraction to detect analogies more effectively. As discussed in \autoref{method:abstraction}, abstraction can occur at multiple levels, and we defined four distinct layers in line with prior theoretical work on framing, each capturing a different aspect of meaning (Conceptual, Evaluative, Narrative Arc, and Stage). Still, designing abstraction levels for narratives that are both expressive and reliable remains a challenge, and our four levels serve as a first operational step towards that goal. 
Future abstraction designs should incorporate higher layers of meaning, such as emotions and morality. While we made an initial attempt to model emotion through the Evaluative Abstraction, future work can explore how best to incorporate a principled representation of emotions~\citep{thagard2001emotional}. Similarly, morality remains underexplored. Understood as culturally grounded solutions to social problems \citep{curry2022moral}, morality plays a central role in storytelling, yet no dedicated abstraction layer captures it in \frameworkName. Theories such as Moral Molecules suggest that complex moral judgments emerge from combinations of simpler atomic concepts \citep{curry2022moral}, resembling our stage abstraction layer. However, a morality-based approach would require first identifying atomic moral constituents before composing higher-order structures.

A related consideration is whether abstractions should be derived from a single story or from both stories jointly. When considering one story at a time, an abstraction valid in isolation may be misaligned for mapping to another story. While our multiple levels of abstraction support this plurality of interpretation to some extent, more principled joint abstraction would require a different design of the mapping and scoring mechanism. Future work can explore cognitive methods for conceptual blending~\citep{fauconnier2003conceptual}, LLM multi-agent debate methods~\citep{liang2024encouraging}, and informal argumentation approaches based on entailment trees~\citep{kassner2023language} for this purpose.

\noindent \textbf{How to perform global mapping robustly?} Our mapping component may still produce incorrect answers even when the extracted information is largely correct, because the similarity function relies on event embeddings that capture surface similarity, and the mapping struggles with different patterns of story analogy (\autoref{sec:error_analysis}). Several directions could address these challenges: replacing the greedy algorithm with a max-flow-based algorithm \citep{yuan2010study} could yield an optimal global mapping, graph edit distance can score the similarity of graph-based inputs, and NLI-based scoring \citep{williams2018broad} appears well-suited to capturing contradictions. In preliminary experiments, NLI scores achieved 0.6 accuracy on the far-near ARN category, surpassing all prior results and supporting our hypothesis that ARN encompasses multiple analogy patterns that structural mapping alone cannot fully capture. Beyond structural mapping, alternative reasoning engines such as Copycat \citep{hofstadter1995copycat} and ACME \citep{holyoak1989analogical} (reviewed in \autoref{sec:related}) may be promising avenues for future work. 

\noindent \textbf{How broad should the analogy benchmarks be, and which patterns are relevant?}
We identified several shortcomings of the two benchmarks, such as reliance on grammatical similarity in the MCQ dataset and the presence of multiple analogy patterns in the ARN dataset (\autoref{sec:error_analysis}). These observations suggest that the definition of story analogy in these benchmarks is unclear and not well-defined. In addition, current datasets lack fine-grained annotations of relational alignments and mappings, making it difficult to evaluate analogical reasoning at a structural level. Together, these limitations highlight the need for improved benchmarks for both evaluation and training. Future work should focus on developing datasets that enforce relational alignment, control for surface cues, and distinguish reasoning processes from outcomes. Inspired by work on story emotion arcs \citep{reagan2016emotional} and abstract moral reasoning \citep{marcuzzo2025morables}, such datasets could combine LLMs with human evaluation to produce detailed relational mappings of narratives, capturing both the sequence of event mappings and the outcome of such mappings. In addition, recent research has shown that fine-tuning LLMs can improve their behavior and reasoning at test time \citep{feuer2025wildchat}. However, no dataset currently exists for analogical reasoning to enable such fine-tuning. A promising direction for future work is to develop such a dataset, fine-tune LLMs on it, and evaluate how their analogical reasoning performance changes.

\section{Conclusions}
We proposed \frameworkName\unskip, a framework for analogical reasoning in narratives that leverages LLMs to extract information from raw stories and applies structural mapping over the resulting representations. To support this process, we introduced four levels of abstraction, grounded in framing theories, that play a central role in enabling robust mapping beyond surface similarity. The evaluation of our approach on two narrative analogy datasets demonstrated that, while LLMs struggle with analogical reasoning when structural mapping is required, combining abstraction with structural mapping yields competitive, and in some cases superior, performance. Additionally, by analyzing different abstraction settings, we find that their effectiveness depends on the level of surface similarity, and no single abstraction works best across all settings. We observed that LLMs are inconsistent across different unit granularities and instructions, making them unreliable for analogical reasoning. Our error analysis identifies where and why failures occur in the \frameworkName pipeline, including in event extraction, abstraction (e.g., causal misinterpretation), mapping (surface-level similarity), and dataset design (multiple analogy patterns in ARN).

We view this work as a first step toward operationalizing cognitive theories of analogical reasoning over raw natural language and systematically exploring their integration with LLMs. Our results suggest that while LLMs alone are unreliable for abstraction and analogical reasoning, their combination with cognitively inspired mapping mechanisms is promising. At the same time, far analogies remain difficult. Future directions for advancing analogical reasoning in AI systems include exploring structured unit representations, integrating alternative cognitive frameworks, enriching the abstraction levels with theories of morality and emotion, and developing analogy datasets with fine-grained mapping annotations.

\section*{Acknowledgements}

The authors thank Emile van Krieken, Gabriella Bollici, Fabian Hoppe, Ting-Chih Chen, Bradley P. Allen, Shaurya Gaur, and Zhivar Sourati for their helpful comments and suggestions.

\bibliographystyle{unsrt}  
\bibliography{references}  

@article{forbus2017extending,
  title={Extending SME to handle large-scale cognitive modeling},
  author={Forbus, Kenneth D and Ferguson, Ronald W and Lovett, Andrew and Gentner, Dedre},
  journal={Cognitive Science},
  volume={41},
  number={5},
  pages={1152--1201},
  year={2017},
  publisher={Wiley Online Library}
}

@inproceedings{liang2024encouraging,
  title={Encouraging divergent thinking in large language models through multi-agent debate},
  author={Liang, Tian and He, Zhiwei and Jiao, Wenxiang and Wang, Xing and Wang, Yan and Wang, Rui and Yang, Yujiu and Shi, Shuming and Tu, Zhaopeng},
  booktitle={Proceedings of the 2024 conference on empirical methods in natural language processing},
  pages={17889--17904},
  year={2024}
}

@book{propp2010morphology,
  title={Morphology of the folk tale},
  author={Propp, Vladimir},
  volume={10},
  year={2010},
  publisher={Univ of TX+ ORM}
}

@article{fauconnier2003conceptual,
  title={Conceptual blending, form and meaning},
  author={Fauconnier, Gilles and Turner, Mark},
  journal={Recherches en communication},
  volume={19},
  pages={57--86},
  year={2003}
}

@inproceedings{kassner2023language,
  title={Language models with rationality},
  author={Kassner, Nora and Tafjord, Oyvind and Sabharwal, Ashish and Richardson, Kyle and Schuetze, Hinrich and Clark, Peter},
  booktitle={Proceedings of the 2023 conference on empirical methods in natural language processing},
  pages={14190--14201},
  year={2023}
}

@article{liu2020extracting,
  title={Extracting events and their relations from texts: A survey on recent research progress and challenges},
  author={Liu, Kang and Chen, Yubo and Liu, Jian and Zuo, Xinyu and Zhao, Jun},
  journal={AI Open},
  volume={1},
  pages={22--39},
  year={2020},
  publisher={Elsevier}
}

@inproceedings{lee2021maximum,
    title = "Maximum {B}ayes {S}match Ensemble Distillation for {AMR} Parsing",
    author = "Lee, Young-Suk  and
      Astudillo, Ram{\'o}n  and
      Thanh Lam, Hoang  and
      Naseem, Tahira  and
      Florian, Radu  and
      Roukos, Salim",
    editor = "Carpuat, Marine  and
      de Marneffe, Marie-Catherine  and
      Meza Ruiz, Ivan Vladimir",
    booktitle = "Proceedings of the 2022 Conference of the North American Chapter of the Association for Computational Linguistics: Human Language Technologies",
    month = jul,
    year = "2022",
    address = "Seattle, United States",
    publisher = "Association for Computational Linguistics",
    url = "https://aclanthology.org/2022.naacl-main.393",
    doi = "10.18653/v1/2022.naacl-main.393",
    pages = "5379--5392",
}

@Book{jm3,
  author =       "Daniel Jurafsky and James H. Martin",
  title =        "Speech and Language Processing: An Introduction to Natural Language Processing, Computational Linguistics, and Speech Recognition, with Language Models",
  year =         "2025",
  url = {https://web.stanford.edu/~jurafsky/slp3/},
  note = "Online manuscript released August 24, 2025",
  edition =         "3rd",
  }

@inproceedings{li-etal-2023-open,
    title = "Open-Domain Hierarchical Event Schema Induction by Incremental Prompting and Verification",
    author = "Li, Sha  and
      Zhao, Ruining  and
      Li, Manling  and
      Ji, Heng  and
      Callison-Burch, Chris  and
      Han, Jiawei",
    editor = "Rogers, Anna  and
      Boyd-Graber, Jordan  and
      Okazaki, Naoaki",
    booktitle = "Proceedings of the 61st Annual Meeting of the Association for Computational Linguistics (Volume 1: Long Papers)",
    month = jul,
    year = "2023",
    address = "Toronto, Canada",
    publisher = "Association for Computational Linguistics",
    url = "https://aclanthology.org/2023.acl-long.312/",
    doi = "10.18653/v1/2023.acl-long.312",
}

@article{mu2025netprompt,
  title={NetPrompt: Neural Network Prompting Enhances Event Extraction in Large Language Models},
  author={Mu, Lin and Cheng, Yide and Shen, Jun and Zhang, Yiwen and Zhong, Hong},
  journal={IEEE Transactions on Big Data},
  year={2025},
  publisher={IEEE}
}

@inproceedings{hsu-etal-2023-ampere,
    title = "{AMPERE}: {AMR}-Aware Prefix for Generation-Based Event Argument Extraction Model",
    author = "Hsu, I-Hung  and
      Xie, Zhiyu  and
      Huang, Kuan-Hao  and
      Natarajan, Prem  and
      Peng, Nanyun",
    editor = "Rogers, Anna  and
      Boyd-Graber, Jordan  and
      Okazaki, Naoaki",
    booktitle = "Proceedings of the 61st Annual Meeting of the Association for Computational Linguistics (Volume 1: Long Papers)",
    month = jul,
    year = "2023",
    address = "Toronto, Canada",
    publisher = "Association for Computational Linguistics",
    url = "https://aclanthology.org/2023.acl-long.615/",
    doi = "10.18653/v1/2023.acl-long.615",
    pages = "10976--10993"
}

@inproceedings{mostafazadeh2016corpus,
  title={A corpus and cloze evaluation for deeper understanding of commonsense stories},
  author={Mostafazadeh, Nasrin and Chambers, Nathanael and He, Xiaodong and Parikh, Devi and Batra, Dhruv and Vanderwende, Lucy and Kohli, Pushmeet and Allen, James},
  booktitle={Proceedings of the 2016 Conference of the North American Chapter of the Association for Computational Linguistics: Human Language Technologies},
  pages={839--849},
  year={2016}
}

@article{yang2025qwen3,
  title={Qwen3 technical report},
  author={Yang, An and Li, Anfeng and Yang, Baosong and Zhang, Beichen and Hui, Binyuan and Zheng, Bo and Yu, Bowen and Gao, Chang and Huang, Chengen and Lv, Chenxu and others},
  journal={arXiv preprint arXiv:2505.09388},
  year={2025}
}

@article{gentner2017analogy,
  title={Analogy and abstraction},
  author={Gentner, Dedre and Hoyos, Christian},
  journal={Topics in cognitive science},
  volume={9},
  number={3},
  pages={672--693},
  year={2017},
  publisher={Wiley Online Library}
}

@article{barsalou2003abstraction,
    author = {Barsalou, Lawrence W.},
    title = {Abstraction in perceptual symbol systems},
    journal = {Philosophical Transactions of the Royal Society B: Biological Sciences},
    volume = {358},
    number = {1435},
    pages = {1177-1187},
    year = {2003},
    month = {07},
    issn = {0962-8436},
    doi = {10.1098/rstb.2003.1319},
    url = {https://doi.org/10.1098/rstb.2003.1319},
    eprint = {https://royalsocietypublishing.org/rstb/article-pdf/358/1435/1177/87206/rstb.2003.1319.pdf},
}

@article{bolognesi2020abstraction,
  title={On abstraction: decoupling conceptual concreteness and categorical specificity},
  author={Bolognesi, Marianna and Burgers, Christian and Caselli, Tommaso},
  journal={Cognitive Processing},
  volume={21},
  number={3},
  pages={365--381},
  year={2020},
  publisher={Springer}
}

@article{das2014frame,
  title={Frame-semantic parsing},
  author={Das, Dipanjan and Chen, Desai and Martins, Andr{\'e} FT and Schneider, Nathan and Smith, Noah A},
  journal={Computational linguistics},
  volume={40},
  number={1},
  pages={9--56},
  year={2014},
  publisher={MIT Press One Rogers Street, Cambridge, MA 02142-1209, USA journals-info~…}
}

@article{palmer2005proposition,
  title={The proposition bank: An annotated corpus of semantic roles},
  author={Palmer, Martha and Gildea, Daniel and Kingsbury, Paul},
  journal={Computational linguistics},
  volume={31},
  number={1},
  pages={71--106},
  year={2005},
  publisher={MIT press One Rogers Street, Cambridge, MA 02142-1209, USA journals-info~…}
}

@inproceedings{banarescu2013abstract,
  title={Abstract meaning representation for sembanking},
  author={Banarescu, Laura and Bonial, Claire and Cai, Shu and Georgescu, Madalina and Griffitt, Kira and Hermjakob, Ulf and Knight, Kevin and Koehn, Philipp and Palmer, Martha and Schneider, Nathan},
  booktitle={Proceedings of the 7th linguistic annotation workshop and interoperability with discourse},
  pages={178--186},
  year={2013}
}

@article{sullivan2023three,
  title={Three levels of framing},
  author={Sullivan, Karen},
  journal={Wiley Interdisciplinary Reviews: Cognitive Science},
  volume={14},
  number={5},
  pages={e1651},
  year={2023},
  publisher={Wiley Online Library}
}

@inproceedings{jacob-etal-2023-fame,
    title = "{FAME}: Flexible, Scalable Analogy Mappings Engine",
    author = "Jacob, Shahar  and
      Shani, Chen  and
      Shahaf, Dafna",
    editor = "Bouamor, Houda  and
      Pino, Juan  and
      Bali, Kalika",
    booktitle = "Proceedings of the 2023 Conference on Empirical Methods in Natural Language Processing",
    month = dec,
    year = "2023",
    address = "Singapore",
    publisher = "Association for Computational Linguistics",
    url = "https://aclanthology.org/2023.emnlp-main.1023/",
    doi = "10.18653/v1/2023.emnlp-main.1023",
    pages = "16426--16442"
}

@inproceedings{jiayang-etal-2023-storyanalogy,
    title = "{S}tory{A}nalogy: Deriving Story-level Analogies from Large Language Models to Unlock Analogical Understanding",
    author = "Jiayang, Cheng  and
      Qiu, Lin  and
      Chan, Tsz  and
      Fang, Tianqing  and
      Wang, Weiqi  and
      Chan, Chunkit  and
      Ru, Dongyu  and
      Guo, Qipeng  and
      Zhang, Hongming  and
      Song, Yangqiu  and
      Zhang, Yue  and
      Zhang, Zheng",
    editor = "Bouamor, Houda  and
      Pino, Juan  and
      Bali, Kalika",
    booktitle = "Proceedings of the 2023 Conference on Empirical Methods in Natural Language Processing",
    month = dec,
    year = "2023",
    address = "Singapore",
    publisher = "Association for Computational Linguistics",
    url = "https://aclanthology.org/2023.emnlp-main.706/",
    doi = "10.18653/v1/2023.emnlp-main.706",
    pages = "11518--11537"
}

@article{sourati-etal-2024-arn,
    title = "{ARN}: Analogical Reasoning on Narratives",
    author = "Sourati, Zhivar  and
      Ilievski, Filip  and
      Sommerauer, Pia  and
      Jiang, Yifan",
    journal = "Transactions of the Association for Computational Linguistics",
    volume = "12",
    year = "2024",
    address = "Cambridge, MA",
    publisher = "MIT Press",
    url = "https://aclanthology.org/2024.tacl-1.59/",
    doi = "10.1162/tacl_a_00688",
    pages = "1063--1086"
}

@article{grattafiori2024llama,
  title={The llama 3 herd of models},
  author={Grattafiori, Aaron and Dubey, Abhimanyu and Jauhri, Abhinav and Pandey, Abhinav and Kadian, Abhishek and Al-Dahle, Ahmad and Letman, Aiesha and Mathur, Akhil and Schelten, Alan and Vaughan, Alex and others},
  journal={arXiv preprint arXiv:2407.21783},
  year={2024}
}

@inproceedings{baker1998berkeley,
  title={The berkeley framenet project},
  author={Baker, Collin F and Fillmore, Charles J and Lowe, John B},
  booktitle={COLING 1998 Volume 1: The 17th International Conference on Computational Linguistics},
  year={1998}
}

@inproceedings{williams2018broad,
  title={A broad-coverage challenge corpus for sentence understanding through inference},
  author={Williams, Adina and Nangia, Nikita and Bowman, Samuel},
  booktitle={Proceedings of the 2018 conference of the North American chapter of the association for computational linguistics: human language technologies, volume 1 (long papers)},
  pages={1112--1122},
  year={2018}
}

@inproceedings{wein2024survey,
  title={A survey of AMR applications},
  author={Wein, Shira and Opitz, Juri},
  booktitle={Proceedings of the 2024 Conference on Empirical Methods in Natural Language Processing},
  pages={6856--6875},
  year={2024}
}

@article{jiang2025comprehensive,
  title={A Comprehensive Survey of Subgraph Matching:[Experiments \& Analysis]},
  author={Jiang, Haolin and Pandey, Santosh and Liu, Hang},
  journal={Proceedings of the ACM on Management of Data},
  volume={3},
  number={6},
  pages={1--30},
  year={2025},
  publisher={ACM New York, NY, USA}
}

@article{zhu2024survey,
  title={A survey: knowledge graph entity alignment research based on graph embedding},
  author={Zhu, Beibei and Wang, Ruolin and Wang, Junyi and Shao, Fei and Wang, Kerun},
  journal={Artificial Intelligence Review},
  volume={57},
  number={9},
  pages={229},
  year={2024},
  publisher={Springer}
}

@article{jain2024graph,
  title={Graph edit distance with general costs using neural set divergence},
  author={Jain, Eeshaan and Roy, Indradyumna and Meher, Saswat and Chakrabarti, Soumen and De, Abir},
  journal={Advances in Neural Information Processing Systems},
  volume={37},
  pages={73399--73438},
  year={2024}
}

@article{wang2024guiding,
  title={Guiding and diversifying LLM-based story generation via answer set programming},
  author={Wang, Phoebe J and Kreminski, Max},
  journal={arXiv preprint arXiv:2406.00554},
  year={2024}
}

@article{santana2025question,
  title={Question Answering with LLMs and Learning from Answer Sets},
  author={Santana, Manuel Alejandro Borroto and Gallagher, Katie and Ielo, Antonio and Kareem, Irfan and Ricca, Francesco and Russo, Alessandra},
  journal={Theory and Practice of Logic Programming},
  pages={1--25},
  year={2025},
  publisher={Cambridge University Press}
}

@article{thagard2001emotional,
  title={Emotional Analogies and Analogical},
  author={Thagard, Paul and Shelley, Cameron},
  journal={The analogical mind: Perspectives from cognitive science},
  pages={335},
  year={2001},
  publisher={MIT Press}
}

@article{curry2022moral,
  title={Moral molecules: Morality as a combinatorial system},
  author={Curry, Oliver Scott and Alfano, Mark and Brandt, Mark J and Pelican, Christine},
  journal={Review of Philosophy and Psychology},
  volume={13},
  number={4},
  pages={1039--1058},
  year={2022},
  publisher={Springer}
}

@article{lee2025curious,
  title={The Curious Case of Analogies: Investigating Analogical Reasoning in Large Language Models},
  author={Lee, Taewhoo and Song, Minju and Yoon, Chanwoong and Park, Jungwoo and Kang, Jaewoo},
  journal={arXiv preprint arXiv:2511.20344},
  year={2025}
}

@article{hofstadter1995copycat,
  title={The copycat project: A model of mental fluidity and analogy-making},
  author={Hofstadter, Douglas R and Mitchell, Melanie},
  journal={Advances in connectionist and neural computation theory},
  volume={2},
  number={205-267},
  pages={2--3},
  year={1995},
  publisher={Ablex Norwood, NJ}
}

@article{holyoak1989analogical,
  title={Analogical mapping by constraint satisfaction},
  author={Holyoak, Keith J and Thagard, Paul},
  journal={Cognitive science},
  volume={13},
  number={3},
  pages={295--355},
  year={1989},
  publisher={Wiley Online Library}
}

@article{wei2022chain,
  title={Chain-of-thought prompting elicits reasoning in large language models},
  author={Wei, Jason and Wang, Xuezhi and Schuurmans, Dale and Bosma, Maarten and Xia, Fei and Chi, Ed and Le, Quoc V and Zhou, Denny and others},
  journal={Advances in neural information processing systems},
  volume={35},
  pages={24824--24837},
  year={2022}
}

@incollection{GENTNER2012130,
title = {Analogical Reasoning},
editor = {V.S. Ramachandran},
booktitle = {Encyclopedia of Human Behavior (Second Edition)},
publisher = {Academic Press},
edition = {Second Edition},
address = {San Diego},
pages = {130-136},
year = {2012},
isbn = {978-0-08-096180-4},
doi = {https://doi.org/10.1016/B978-0-12-375000-6.00022-7},
url = {https://www.sciencedirect.com/science/article/pii/B9780123750006000227},
author = {D. Gentner and L. Smith}
}

@inproceedings{guo2024nutframe,
  title={NutFrame: Frame-based Conceptual Structure Induction with LLMs},
  author={Guo, Shaoru and Chen, Yubo and Liu, Kang and Li, Ru and Zhao, Jun},
  booktitle={Proceedings of the 2024 Joint International Conference on Computational Linguistics, Language Resources and Evaluation (LREC-COLING 2024)},
  pages={12330--12335},
  year={2024}
}

@article{yang2022towards,
  title={Towards fine-grained causal reasoning and qa},
  author={Yang, Linyi and Wang, Zhen and Wu, Yuxiang and Yang, Jie and Zhang, Yue},
  journal={arXiv preprint arXiv:2204.07408},
  year={2022}
}

@inproceedings{sun2024event,
  title={Event causality is key to computational story understanding},
  author={Sun, Yidan and Chao, Qin and Li, Boyang},
  booktitle={Proceedings of the 2024 Conference of the North American Chapter of the Association for Computational Linguistics: Human Language Technologies (Volume 1: Long Papers)},
  pages={3493--3511},
  year={2024}
}

@inproceedings{hu2025large,
  title={Large Language Model-Based Event Relation Extraction with Rationales},
  author={Hu, Zhilei and Li, Zixuan and Jin, Xiaolong and Bai, Long and Guo, Jiafeng and Cheng, Xueqi},
  booktitle={Proceedings of the 31st International Conference on Computational Linguistics},
  pages={7484--7496},
  year={2025}
}

@inproceedings{wei2024llms,
  title={Are LLMs Good Annotators for Discourse-level Event Relation Extraction?},
  author={Wei, Kangda and Gautam, Aayush and Huang, Ruihong},
  booktitle={Findings of the Association for Computational Linguistics: EMNLP 2024},
  pages={1--19},
  year={2024}
}

@article{boyd2020narrative,
  title={The narrative arc: Revealing core narrative structures through text analysis},
  author={Boyd, Ryan L and Blackburn, Kate G and Pennebaker, James W},
  journal={Science advances},
  volume={6},
  number={32},
  pages={eaba2196},
  year={2020},
  publisher={American Association for the Advancement of Science}
}

@inproceedings{errica2025did,
  title={What did i do wrong? quantifying llms’ sensitivity and consistency to prompt engineering},
  author={Errica, Federico and Sanvito, Davide and Siracusano, Giuseppe and Bifulco, Roberto},
  booktitle={Proceedings of the 2025 Conference of the Nations of the Americas Chapter of the Association for Computational Linguistics: Human Language Technologies (Volume 1: Long Papers)},
  pages={1543--1558},
  year={2025}
}

@inproceedings{steck2024cosine,
  title={Is cosine-similarity of embeddings really about similarity?},
  author={Steck, Harald and Ekanadham, Chaitanya and Kallus, Nathan},
  booktitle={Companion Proceedings of the ACM Web Conference 2024},
  pages={887--890},
  year={2024}
}

@book{holyoak1996mental,
  title={Mental leaps: Analogy in creative thought},
  author={Holyoak, Keith J and Thagard, Paul},
  year={1996},
  publisher={MIT press}
}

@article{penn2008darwin,
  title={Darwin's mistake: Explaining the discontinuity between human and nonhuman minds},
  author={Penn, Derek C and Holyoak, Keith J and Povinelli, Daniel J},
  journal={Behavioral and brain sciences},
  volume={31},
  number={2},
  pages={109--130},
  year={2008},
  publisher={Cambridge University Press}
}

@article{hofstadter2001analogy,
  title={Analogy as the core of cognition},
  author={Hofstadter, Douglas R and others},
  journal={The analogical mind: Perspectives from cognitive science},
  pages={499--538},
  year={2001}
}

@inproceedings{lewis2024using,
  title={Using Counterfactual Tasks to Evaluate the Generality of Analogical Reasoning in Large Language Models},
  author={Lewis, Martha and Mitchell, Melanie},
  booktitle={Proceedings of the Annual Meeting of the Cognitive Science Society},
  volume={46},
  year={2024}
}

@article{opielka2025analogical,
  title={Analogical reasoning inside large language models: Concept vectors and the limits of abstraction},
  author={Opie{\l}ka, Gustaw and Rosenbusch, Hannes and Stevenson, Claire E},
  journal={arXiv preprint arXiv:2503.03666},
  year={2025}
}

@article{gentner1983structure,
  title={Structure-mapping: A theoretical framework for analogy},
  author={Gentner, Dedre},
  journal={Cognitive science},
  volume={7},
  number={2},
  pages={155--170},
  year={1983},
  publisher={Elsevier}
}

@article{gentner2025structure,
  title={The Structure-Mapping Engine: A Multidecade Interaction Between Psychology and Artificial Intelligence},
  author={Gentner, Dedre and Forbus, Kenneth},
  journal={Current Directions in Psychological Science},
  pages={09637214251395678},
  year={2025},
  publisher={SAGE Publications Sage CA: Los Angeles, CA}
}

@article{mikolov2013efficient,
  title={Efficient estimation of word representations in vector space},
  author={Mikolov, Tomas and Chen, Kai and Corrado, Greg and Dean, Jeffrey},
  journal={arXiv preprint arXiv:1301.3781},
  year={2013}
}

@inproceedings{ushio2021bert,
  title={BERT is to NLP what AlexNet is to CV: Can pre-trained language models identify analogies?},
  author={Ushio, Asahi and Anke, Luis Espinosa and Schockaert, Steven and Camacho-Collados, Jose},
  booktitle={Proceedings of the 59th Annual Meeting of the Association for Computational Linguistics and the 11th International Joint Conference on Natural Language Processing (Volume 1: Long Papers)},
  pages={3609--3624},
  year={2021}
}

@article{webb2023emergent,
  title={Emergent analogical reasoning in large language models},
  author={Webb, Taylor and Holyoak, Keith J and Lu, Hongjing},
  journal={Nature Human Behaviour},
  volume={7},
  number={9},
  pages={1526--1541},
  year={2023},
  publisher={Nature Publishing Group UK London}
}

@article{stevenson2024can,
  title={Can Large Language Models generalize analogy solving like children can?},
  author={Stevenson, Claire E and Pafford, Alexandra and van der Maas, Han LJ and Mitchell, Melanie},
  journal={arXiv preprint arXiv:2411.02348},
  year={2024}
}

@article{falkenhainer1989structure,
  title={The structure-mapping engine: Algorithm and examples},
  author={Falkenhainer, Brian and Forbus, Kenneth D and Gentner, Dedre},
  journal={Artificial intelligence},
  volume={41},
  number={1},
  pages={1--63},
  year={1989},
  publisher={Elsevier}
}

@article{hummel1997distributed,
  title={Distributed representations of structure: A theory of analogical access and mapping.},
  author={Hummel, John E and Holyoak, Keith J},
  journal={Psychological review},
  volume={104},
  number={3},
  pages={427},
  year={1997},
  publisher={American Psychological Association}
}

@inproceedings{pennington2014glove,
  title={Glove: Global vectors for word representation},
  author={Pennington, Jeffrey and Socher, Richard and Manning, Christopher D},
  booktitle={Proceedings of the 2014 conference on empirical methods in natural language processing (EMNLP)},
  pages={1532--1543},
  year={2014}
}

@inproceedings{rogers2017too,
  title={The (too many) problems of analogical reasoning with word vectors},
  author={Rogers, Anna and Drozd, Aleksandr and Li, Bofang},
  booktitle={Proceedings of the 6th joint conference on lexical and computational semantics (* SEM 2017)},
  pages={135--148},
  year={2017}
}

@inproceedings{tian-etal-2024-large-language,
    title = "Are Large Language Models Capable of Generating Human-Level Narratives?",
    author = "Tian, Yufei  and
      Huang, Tenghao  and
      Liu, Miri  and
      Jiang, Derek  and
      Spangher, Alexander  and
      Chen, Muhao  and
      May, Jonathan  and
      Peng, Nanyun",
    editor = "Al-Onaizan, Yaser  and
      Bansal, Mohit  and
      Chen, Yun-Nung",
    booktitle = "Proceedings of the 2024 Conference on Empirical Methods in Natural Language Processing",
    month = nov,
    year = "2024",
    address = "Miami, Florida, USA",
    publisher = "Association for Computational Linguistics",
    url = "https://aclanthology.org/2024.emnlp-main.978/",
    doi = "10.18653/v1/2024.emnlp-main.978",
    pages = "17659--17681"
}

@inproceedings{sun-etal-2024-event,
    title = "Event Causality Is Key to Computational Story Understanding",
    author = "Sun, Yidan  and
      Chao, Qin  and
      Li, Boyang",
    editor = "Duh, Kevin  and
      Gomez, Helena  and
      Bethard, Steven",
    booktitle = "Proceedings of the 2024 Conference of the North American Chapter of the Association for Computational Linguistics: Human Language Technologies (Volume 1: Long Papers)",
    month = jun,
    year = "2024",
    address = "Mexico City, Mexico",
    publisher = "Association for Computational Linguistics",
    url = "https://aclanthology.org/2024.naacl-long.191/",
    doi = "10.18653/v1/2024.naacl-long.191",
    pages = "3493--3511"
}

@inproceedings{hu-etal-2025-large,
    title = "Large Language Model-Based Event Relation Extraction with Rationales",
    author = "Hu, Zhilei  and
      Li, Zixuan  and
      Jin, Xiaolong  and
      Bai, Long  and
      Guo, Jiafeng  and
      Cheng, Xueqi",
    editor = "Rambow, Owen  and
      Wanner, Leo  and
      Apidianaki, Marianna  and
      Al-Khalifa, Hend  and
      Eugenio, Barbara Di  and
      Schockaert, Steven",
    booktitle = "Proceedings of the 31st International Conference on Computational Linguistics",
    month = jan,
    year = "2025",
    address = "Abu Dhabi, UAE",
    publisher = "Association for Computational Linguistics",
    url = "https://aclanthology.org/2025.coling-main.500/",
    pages = "7484--7496"
}

@inproceedings{yuan-etal-2023-zero,
    title = "Zero-shot Temporal Relation Extraction with {C}hat{GPT}",
    author = "Yuan, Chenhan  and
      Xie, Qianqian  and
      Ananiadou, Sophia",
    editor = "Demner-fushman, Dina  and
      Ananiadou, Sophia  and
      Cohen, Kevin",
    booktitle = "Proceedings of the 22nd Workshop on Biomedical Natural Language Processing and BioNLP Shared Tasks",
    month = jul,
    year = "2023",
    address = "Toronto, Canada",
    publisher = "Association for Computational Linguistics",
    url = "https://aclanthology.org/2023.bionlp-1.7/",
    doi = "10.18653/v1/2023.bionlp-1.7",
    pages = "92--102"
}

@inproceedings{arabshahi2021conversational,
  title={Conversational Multi-Hop Reasoning with Neural Commonsense Knowledge and Symbolic Logic Rules},
  author={Arabshahi, Forough and Lee, Jennifer and Bosselut, Antoine and Choi, Yejin and Mitchell, Tom},
  booktitle={Proceedings of the 2021 Conference on Empirical Methods in Natural Language Processing},
  pages={7404--7418},
  year={2021}
}

@article{wang2020minilm,
  title={Minilm: Deep self-attention distillation for task-agnostic compression of pre-trained transformers},
  author={Wang, Wenhui and Wei, Furu and Dong, Li and Bao, Hangbo and Yang, Nan and Zhou, Ming},
  journal={Advances in neural information processing systems},
  volume={33},
  pages={5776--5788},
  year={2020}
}

@inproceedings{reimers2019sentence,
  title={Sentence-BERT: Sentence Embeddings using Siamese BERT-Networks},
  author={Reimers, Nils and Gurevych, Iryna},
  booktitle={Proceedings of the 2019 Conference on Empirical Methods in Natural Language Processing and the 9th International Joint Conference on Natural Language Processing (EMNLP-IJCNLP)},
  pages={3982--3992},
  year={2019}
}

@inproceedings{kwon2023efficient,
  title={Efficient memory management for large language model serving with pagedattention},
  author={Kwon, Woosuk and Li, Zhuohan and Zhuang, Siyuan and Sheng, Ying and Zheng, Lianmin and Yu, Cody Hao and Gonzalez, Joseph and Zhang, Hao and Stoica, Ion},
  booktitle={Proceedings of the 29th symposium on operating systems principles},
  pages={611--626},
  year={2023}
}

@inproceedings{yuan2010study,
  title={A study on continuous max-flow and min-cut approaches},
  author={Yuan, Jing and Bae, Egil and Tai, Xue-Cheng},
  booktitle={2010 ieee computer society conference on computer vision and pattern recognition},
  pages={2217--2224},
  year={2010},
  organization={IEEE}
}

@article{bal2016medium,
  title={A medium-scale distributed system for computer science research: Infrastructure for the long term},
  author={Bal, Henri and Epema, Dick and De Laat, Cees and Van Nieuwpoort, Rob and Romein, John and Seinstra, Frank and Snoek, Cees and Wijshoff, Harry},
  journal={Computer},
  volume={49},
  number={5},
  pages={54--63},
  year={2016},
  publisher={IEEE}
}

@article{popova2014narrativity,
  title={Narrativity and enaction: the social nature of literary narrative understanding},
  author={Popova, Yanna B},
  journal={Frontiers in psychology},
  volume={5},
  pages={895},
  year={2014},
  publisher={Frontiers Media SA}
}

@article{petersen2025modelling,
  title={Modelling Analogies and Analogical Reasoning: Connecting Cognitive Science Theory and NLP Research},
  author={Petersen, Molly R and Stevenson, Claire E and van der Plas, Lonneke},
  journal={arXiv preprint arXiv:2509.09381},
  year={2025}
}

@inproceedings{turney2003combining,
  title={Combining Independent Modules to Solve Multiple-choice Synonym and Analogy Problems},
  author={Turney, Peter D and Bigham, Jeffrey and Littman, Michael L and Shnayder, Victor},
  booktitle={RECENT ADVANCES IN NATURAL LANGUAGE PROCESSING},
  year={2003}
}

@article{alexieva2017processing,
  title={Processing differences between near and far analogies},
  author={Alexieva, Alexandra and Hristova, Penka},
  year={2017}
}

@article{gentner1986systematicity,
  title={Systematicity and surface similarity in the development of analogy},
  author={Gentner, Dedre and Toupin, Cecile},
  journal={Cognitive science},
  volume={10},
  number={3},
  pages={277--300},
  year={1986},
  publisher={Elsevier}
}

@article{lu2022probabilistic,
  title={Probabilistic analogical mapping with semantic relation networks.},
  author={Lu, Hongjing and Ichien, Nicholas and Holyoak, Keith J},
  journal={Psychological review},
  volume={129},
  number={5},
  pages={1078},
  year={2022},
  publisher={American Psychological Association}
}

@article{ilievski2025aligning,
  title={Aligning generalization between humans and machines},
  author={Ilievski, Filip and Hammer, Barbara and van Harmelen, Frank and Paassen, Benjamin and Saralajew, Sascha and Schmid, Ute and Biehl, Michael and Bolognesi, Marianna and Dong, Xin Luna and Gashteovski, Kiril and others},
  journal={Nature Machine Intelligence},
  volume={7},
  number={9},
  pages={1378--1389},
  year={2025},
  publisher={Nature Publishing Group UK London}
}

@article{minsky1974framework,
  title={A framework for representing knowledge},
  author={Minsky, Marvin},
  year={1974}
}

@article{entman1993framing,
  title={Framing: Towards clarification of a fractured paradigm},
  author={Entman, Robert M},
  journal={McQuail's reader in mass communication theory},
  volume={390},
  pages={397},
  year={1993}
}

@inproceedings{aina2018distributional,
  title={A distributional study of negated adjectives and antonyms},
  author={Aina, Laura and Bernardi, Raffaella and Fern{\'a}ndez, Raquel},
  booktitle={CEUR Workshop Proceedings},
  volume={2253},
  year={2018},
  organization={CEUR Workshop Proceedings}
}

@inproceedings{nikolaev2023universe,
  title={The universe of utterances according to BERT},
  author={Nikolaev, Dmitry and Pad{\'o}, Sebastian},
  booktitle={Proceedings of the 15th International Conference on Computational Semantics},
  pages={99--105},
  year={2023}
}

@article{reagan2016emotional,
  title={The emotional arcs of stories are dominated by six basic shapes},
  author={Reagan, Andrew J and Mitchell, Lewis and Kiley, Dilan and Danforth, Christopher M and Dodds, Peter Sheridan},
  journal={EPJ data science},
  volume={5},
  number={1},
  pages={31},
  year={2016},
  publisher={Springer}
}

@inproceedings{marcuzzo2025morables,
  title={MORABLES: A Benchmark for Assessing Abstract Moral Reasoning in LLMs with Fables},
  author={Marcuzzo, Matteo and Zangari, Alessandro and Albarelli, Andrea and Camacho-Collados, Jose and Pilehvar, Mohammad Taher},
  booktitle={Proceedings of the 2025 Conference on Empirical Methods in Natural Language Processing},
  pages={27715--27739},
  year={2025}
}

@inproceedings{feuer2025wildchat,
  title={WildChat-50M: A Deep Dive Into the Role of Synthetic Data in Post-Training},
  author={Feuer, Benjamin and Hegde, Chinmay},
  booktitle={International Conference on Machine Learning},
  pages={17100--17130},
  year={2025},
  organization={PMLR}
}

\appendix

\setcounter{figure}{0}
\setcounter{table}{0}
\setcounter{equation}{0}

\renewcommand{\thefigure}{\thesection.\arabic{figure}}
\renewcommand{\thetable}{\thesection.\arabic{table}}
\renewcommand{\theequation}{\thesection.\arabic{equation}}

\section{Complementary Results}
\begin{figure*}
    \centering 
    \includegraphics[width=\linewidth]{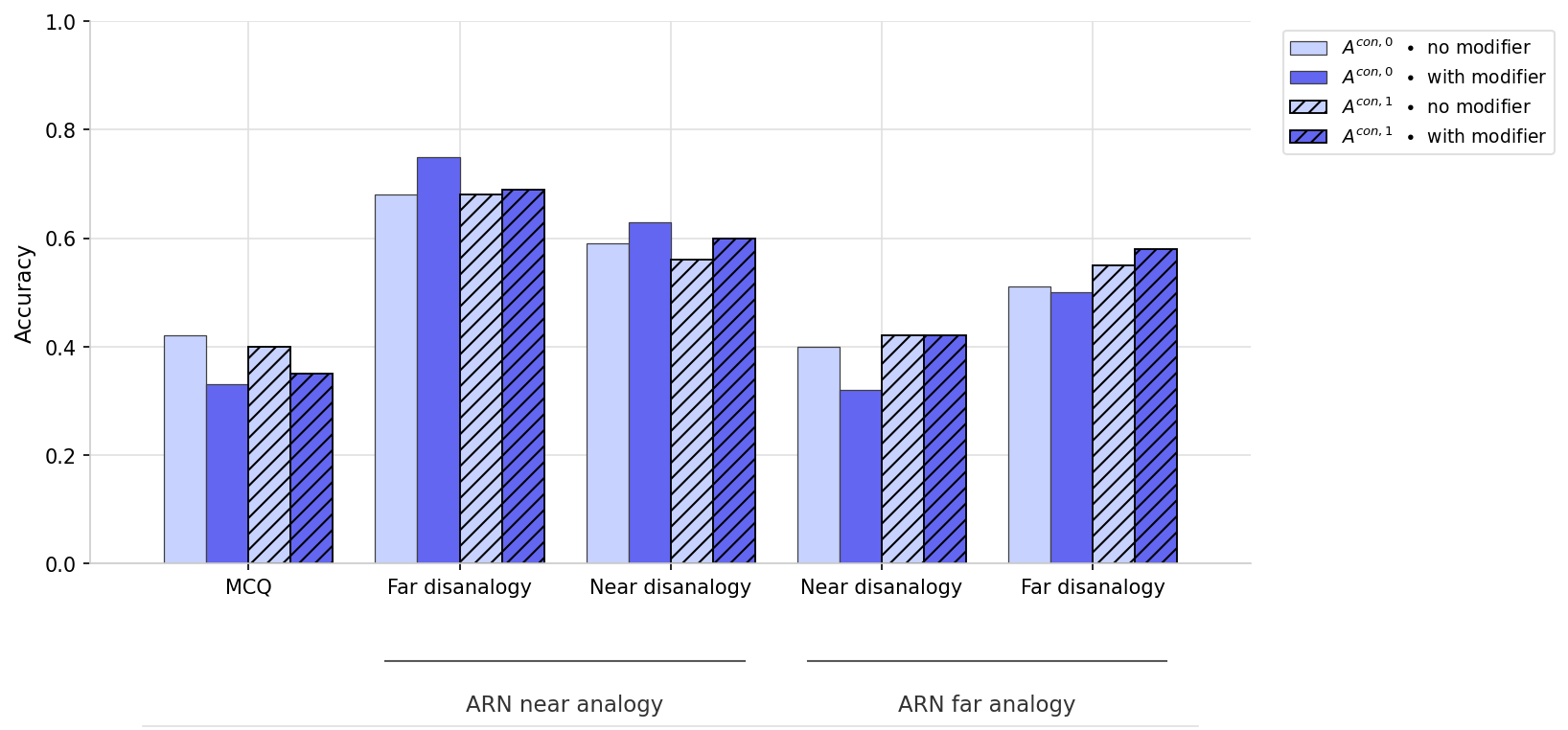} 
    \caption{Effect of hierarchical conceptual abstraction levels for Llama, with and without modifiers.}
    \label{fig:cons-setting-llama} 
\end{figure*}

\begin{figure*} 
    \centering 
    \includegraphics[width=0.7\linewidth]{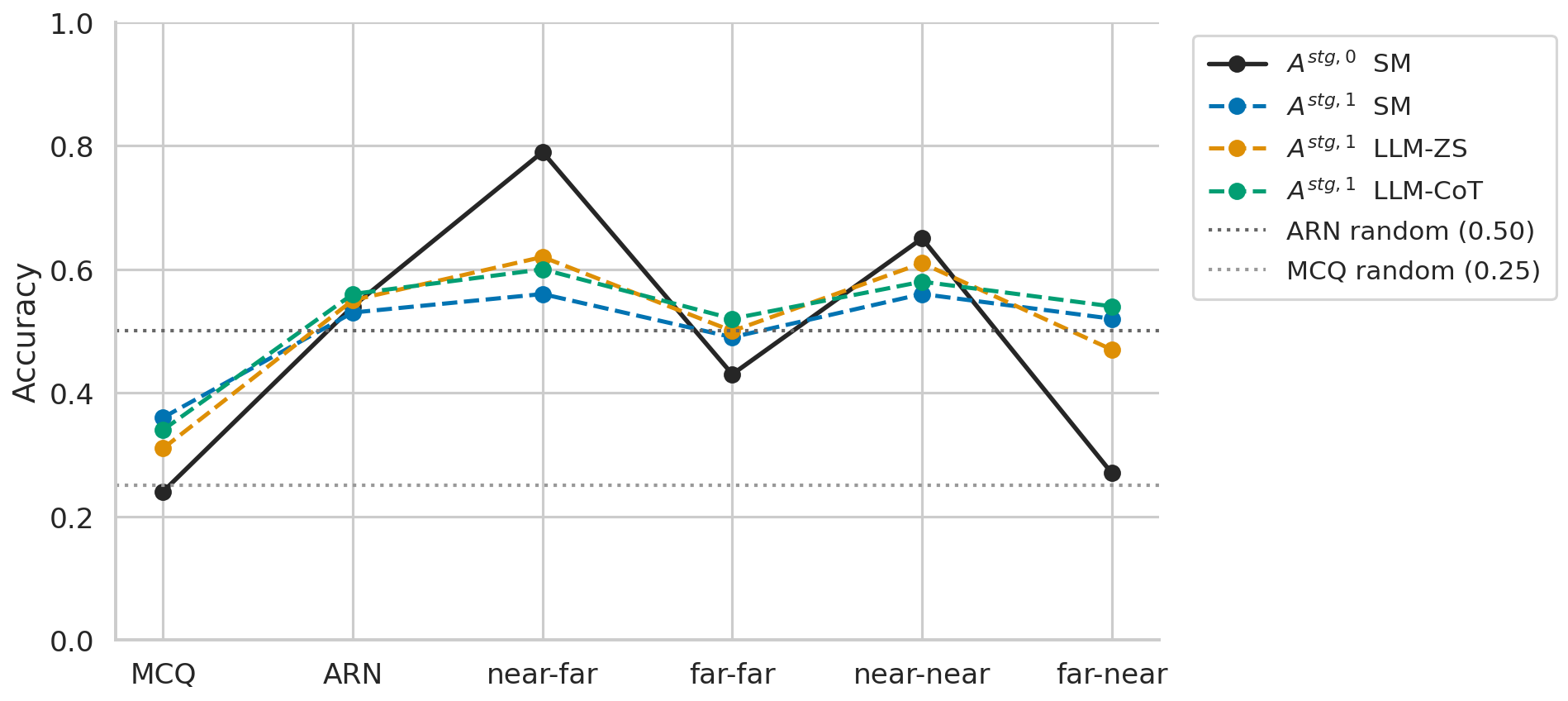} 
    \caption{Effect of hierarchical stage abstraction levels with different settings for Llama.}
    \label{fig:stg-setting-llama} 
\end{figure*}

\begin{figure*} 
    \centering 
    \includegraphics[width=\linewidth]{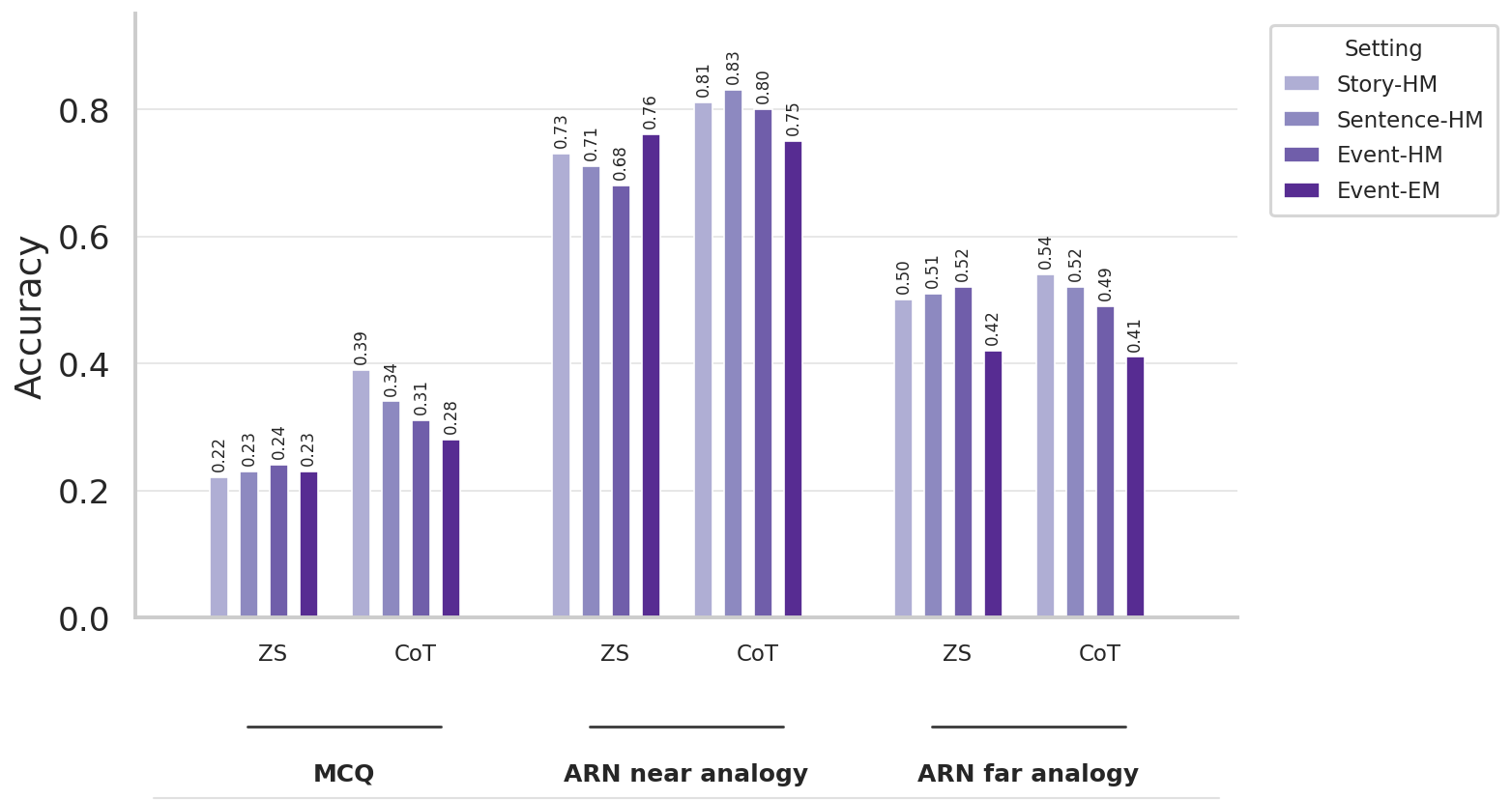} 
    \caption{The result for using different input and mapping instructions with Llama for analogical reasoning.}
    \label{fig:consistent-llama} 
\end{figure*}

In this section, we present complementary results to our main findings. While the main results report ablation studies for the Qwen model, here we provide the corresponding results for Llama. \autoref{fig:cons-setting-llama} shows how performance changes with different conceptual abstraction levels, and \autoref{fig:stg-setting-llama} shows the effect of different stage abstraction levels. \autoref{fig:consistent-llama} further illustrates inconsistencies across input units and prompt settings. We also include two example tables. \autoref{tab:qualitive-example} shows how the Qwen model’s reasoning changes with two different inputs. In addition, \autoref{tab:patterns} illustrates three patterns with examples between the base story and the correct target in analogical reasoning: the first can be captured by structural similarity, while the other two cannot.

\begin{table}[h]
\caption{The model (Qwen) produces different reasoning depending on the input units. In both cases, the model recognizes that the story is the opposite of the base story, but in one case it rejects it for that reason, while in the other it selects it, arguing that the underlying structure is the same. This illustrates the inconsistency of LLMs in analogical reasoning.}
\label{tab:qualitive-example}
\centering
\scriptsize
\setlength{\tabcolsep}{3.5pt}      % tighter columns (default ~6pt)
\renewcommand{\arraystretch}{1.05} % slightly tighter rows

\begin{tabularx}{\linewidth}{@{} X | X @{}}
\toprule
 \textbf{Full story} & \textbf{Event phrases} \\
\hline
This is the opposite of the base story, so it cannot be the answer. & Although the endings are opposite, the structure is the same, so this is the answer. \\
\hline

\end{tabularx}
\end{table}

\begin{table}[t]
\caption{Illustrating examples of different patterns that lead to the correct analogy in the ARN dataset.
\textbf{P1}: Arc-based similarity; \textbf{P2}: Outcome-based similarity; \textbf{P3}: Contrastive similarity.}
\label{tab:patterns}
\centering
\scriptsize
\setlength{\tabcolsep}{3.5pt}      % tighter columns (default ~6pt)
\renewcommand{\arraystretch}{1.05} % slightly tighter rows

\begin{tabularx}{\linewidth}{@{} p{0.09\linewidth} | X | X @{}}
\toprule
\textbf{Pattern} & \textbf{Story 1} & \textbf{Story 2} \\
\midrule
\textbf{P1} &
Johnny is overwhelmed by work and tight deadlines, but his hard work pays off when his boss recognizes him with a raise and greater choice in future projects. &
After a painful breakup and a period of despair, she reluctantly attends a party where she meets someone new, leading to a lasting relationship that is now two years strong. \\
\midrule
\textbf{P2} &
A former Wi-Fi firmware engineer, forced out of work by illness, turns to low-paid crowdsourced micro work and, after months of struggle, celebrates earning enough in a day to feed his family without draining his remaining savings. &
A wireless firmware engineer enters the freelance gig economy to support his family, works diligently through its challenges, and gradually builds a stable and growing income through consistent effort. \\
\midrule
\textbf{P3} &
A pastor realizes too late that focusing on preaching to others has come at the cost of neglecting his own children, who grew up without his guidance and became unbelievers. &
A devout father regularly brings his children to worship and spiritual activities, passing on his faith and values and teaching them that devotion and helping others begin at home. \\
\bottomrule
\end{tabularx}

\end{table}

\section{Technical Details}

\setcounter{figure}{0}
\setcounter{table}{0}
\setcounter{equation}{0}

\subsection{Prompt Strategies}
We provide brief summaries of two prompts used in our experiments. \autoref{tab:event-prompt} presents the prompt for event extraction, while \autoref{tab:cons-prompt} shows the prompt used for conceptual abstraction, illustrating how the model is guided to derive event-level representations and their corresponding abstractions.

\begin{table}[t]
\caption{Prompt for Event extraction}
\label{tab:event-prompt}
\small
\centering
\hspace*{-3cm}%
\begin{tabular}{p{0.8\linewidth}}
\hline

\#\# Role Assignment \\
You are an annotator who reads short stories and converts them into a **list of event phrases**. \\

---
\\
\#\# Task Definition
\\
For each story, extract every **distinct event** and express it as a **single, concise phrase** following the rules below.
\\
---
\\
\#\# Term Definitions
\\
- **event**: The smallest distinct piece of information that is an action, state, fact, intention, reason, or decision.
\\
- **negation**: Explicit “not” forms (e.g., did not, could not), not contractions.
\\
---
\\
\#\# Guidelines
\\
- **One event per element**; do not combine.
\\
- **Extract every event** so the list fully reconstructs the story.
\\
...
\\
---
\\
\#\# Coverage and Self-Check *(silent; output only JSON array)*
\\
- Each sentence yields one or several event.  
\\
- Separate events for multiple clauses/verbs.  
\\
...
\\
---
\\
\#\# Examples
\\
\#\#\# Example 1 \\
Story:
David noticed he had put on a lot of weight recently....
\\
Output:
\begin{verbatim}
<JSON>
[
  David noticed weight gain,
  ...
]
</JSON>
\end{verbatim}
\\
...
\\

\\ \hline
\end{tabular}
\end{table}

\begin{table}[t]
\caption{Prompt for Conceptual abstraction}
\label{tab:cons-prompt}
\small
\centering
\hspace*{-3cm}%
\vspace*{6cm}%
\begin{tabular}{p{0.8\linewidth}}
\hline

\#\# Role Assignment
\\
You are a frame-extraction assistant. Your job is to assign exactly one concise semantic **Frame** to each phrase, and must follow the schema **[MODIFIER]\_[ROOT]**.\\
---
\\
\#\#\# Task Definition
\\
Use the following specification to construct each **Frame**:
\\
**ROOT**: **Purpose:** Name the core event/state/concept in **one noun-like word**.\\
**MODIFIER**: **Purpose:** Provide an **explanation** for the ROOT in **one word**.\\
... \\
---\\
\#\#\# Rationale Style: For each phrase, write a **short self-talk rationale** in **5–7 sentences**.
...\\
---\\
\#\#\# Output Format\\
Return a single JSON object wrapped in **<JSON> ... </JSON>** tags with a top-level **results** array. For each phrase include:\\
**id**, **original\_phrase**, **rationale**, **frame\_name**\\
---\\
\#\# Examples\\
\#\#\# Example 1 \\
Story:\\
David noticed he had put on a lot of weight recently....\\
Phrases:
\begin{verbatim}
[
  {{ "id":"p1","text":"David noticed weight gain" }},
...
]
\end{verbatim}
\\
Output:\\
<JSON>
\begin{verbatim}
{{
  "results": [
    {{
      "id": "p1", "original_phrase": "David noticed weight gain", 
      "rationale": "This event is about recognizing a change in personal condition....", 
      "frame_name": "HEALTH_AWARENESS"
    }}...]
\end{verbatim}
\\ \hline
\end{tabular}
\end{table}

\subsection{Implementation Errors}

We conducted a detailed analysis to ensure that the outputs of both the LLM baselines and our pipeline followed the expected formats and to identify potential errors. For the LLM baselines performing event mapping, we asked the models to return answers in JSON format for CoT and just one number for ZS. In a small number of cases, the models did not follow the requested format, making it impossible to extract the final answer. Across four experimental settings for each model (zero-shot and CoT on two datasets), such cases were rare. For Qwen, we observed two formatting errors in MCQ zero-shot and one in MCQ CoT, and zero problems in ARN. For Llama, we observed five cases in ARN zero-shot and seven in MCQ CoT, and zero problems in other cases. In these situations, the answer was selected randomly. Because these cases are very few, they have a negligible impact on the final results and do not change the ranking of methods.

For our pipeline, two implementation issues may arise. First, the structural mapping component adopted from the FAME framework constructs mappings as quadruples, which requires at least two units per story. If a story contains only a single unit, no mapping can be formed. Second, multiple targets may occasionally receive the same maximum score, resulting in a random selection. In practice, these cases are also rare. When using event phrases or conceptual abstractions, stories typically contain enough units, and we don't have any problems in most cases; while in some cases, we have just one or two problematic cases, mostly because of getting the equal max score for two targets. The problem arises more frequently with stage abstraction, where several events are merged into a single stage. This occurs mainly in the MCQ dataset, which stories are short; after merging events, some stories contain only one stage. This also explains the lower performance observed with stage abstraction on MCQ in our main results. The issue appears less frequently in ARN due to its longer narratives. Our experiments suggest that alternative scoring strategies can mitigate this limitation. For example, using stage abstractions without local and global mapping and forming the mapping using their arc and scoring them with NLI-based similarity produced promising results, particularly for far analogies, without any problematic cases.

\end{document}